\documentclass[acmtog,nonacm]{acmart}

\usepackage{booktabs}
\usepackage{wrapfig}

\setcitestyle{nosort,square} 

\usepackage{enumitem}

\usepackage[ruled]{algorithm2e} 

\SetAlFnt{\small}
\SetAlCapFnt{\small}
\SetAlCapNameFnt{\small}
\SetAlCapHSkip{0pt}

\setcopyright{none}
\makeatletter 

\newcommand{\Rmnum}[1]{\expandafter\@slowromancap\romannumeral #1@}
\makeatother

\newcommand{\hide}[1]{}

\newcommand{\glc}[1]{{\color{orange}         {}}}

\definecolor{revision}{rgb}{0,0,255}
\newcommand{\revision}[1]{{#1}}

\definecolor{rere}{rgb}{0,255,0}

\newcommand{\rere}[1]{{#1}}
\newcommand{\sub}[1]{\text{{#1}}}

\usepackage{multirow}

\begin{document}
\hypersetup{colorlinks=true, allcolors=blue}

\title{Beyond Pixels: Visual Metaphor Transfer via Schema-Driven Agentic Reasoning}

\author{Yu Xu}
\authornote{Equal contribution.}
\orcid{0000-0003-3459-5455}
\author{Yuxin Zhang}
\authornotemark[1]
\orcid{0000-0001-6433-2678}
\author{Lin Gao}
\orcid{0000-0002-1021-8148}
\email{yu.xu915@gmail.com}
\email{yuxin.zhazel@gmail.com}
\email{gaolin@ict.ac.cn}
\affiliation{%
  \institution{University of Chinese Academy of Sciences}
  \country{China}
}

\author{Oliver Deussen}
\orcid{0000-0001-5803-2185}
\email{oliver.deussen@uni-konstanz.de}
\affiliation{%
  \institution{University of Konstanz}
  \country{Germany}
}

\author{Tong-Yee Lee}
\orcid{0000-0001-6699-2944}
\email{tonylee@mail.ncku.edu.tw}
\affiliation{%
  \institution{National Cheng Kung University}
  \country{Taiwan}
}

\author{Fan Tang}
\orcid{0000-0002-3975-2483}
\email{tfan.108@gmail.com}
\affiliation{%
  \institution{University of Science and Technology Beijing}
  \country{China}
}

\authornote{Corresponding author: Fan Tang.}

\begin{abstract}
A visual metaphor constitutes a high-order form of human creativity, employing cross-domain semantic fusion to transform abstract concepts into impactful visual rhetoric. Despite the remarkable progress of generative AI, existing models remain largely confined to pixel-level instruction alignment and surface-level appearance preservation, failing to capture the underlying abstract logic necessary for genuine metaphorical generation. To bridge this gap, we introduce the task of Visual Metaphor Transfer (VMT), which challenges models to autonomously decouple the "creative essence" from a reference image and re-materialize that abstract logic onto a user-specified target subject. We propose a cognitive-inspired, multi-agent framework that operationalizes Conceptual Blending Theory (CBT) through a novel Schema Grammar ($\mathcal{SG}$). This structured representation decouples relational invariants from specific visual entities, providing a rigorous foundation for cross-domain logic re-instantiation. Our pipeline executes VMT through a collaborative system of specialized agents: a perception agent that distills the reference into a schema, a transfer agent that maintains generic space invariance to discover apt carriers, a generation agent for high-fidelity synthesis and a hierarchical diagnostic agent that mimics a professional critic, performing closed-loop backtracking to identify and rectify errors across abstract logic, component selection, and prompt encoding. Extensive experiments and human evaluations demonstrate that our method significantly outperforms state-of-the-art baselines in metaphor consistency, analogy appropriateness, and visual creativity, paving the way for automated high-impact creative applications in advertising and media. Project page with source code and self-contained skills is  \url{https://yuci-gpt.github.io/Beyond-Pixels/}.
\end{abstract}


\begin{teaserfigure}
  \includegraphics[width=\textwidth]{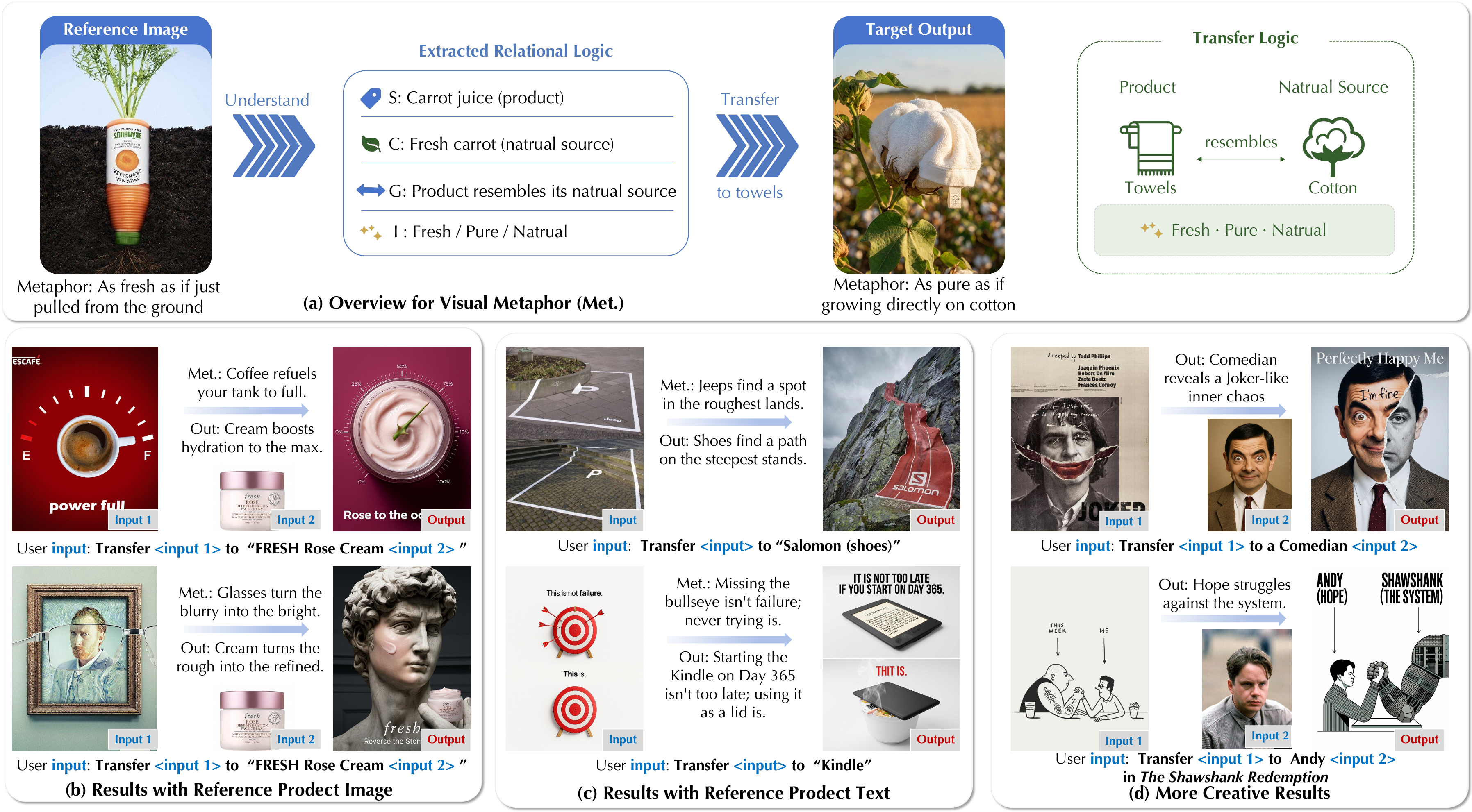}
  \vspace{-7mm}
  \caption{Diverse image metaphor transfer results generated by our framework. For each pair, the left image serves as the \textit{Reference} and the right is the \textit{Generated Result}. 
  Our model demonstrates robust capability across distinct cognitive levels.}
  \label{fig:teaser}
\end{teaserfigure}
\maketitle
\section{Introduction}
\label{sec:intro}
Visual metaphor operates at the upper limits of human creative cognition, where meanings are constructed through the integration of disparate semantic domains, enabling abstract ideas to be conveyed as visually articulated statements with layered, non-literal significance.
Despite the remarkable progress of generative AI, existing text-to-image (T2I)~\cite{rombach2022high, podellsdxl, esser2024scaling, flux, bao2025text} and image-to-image~\cite{mou2025dreamo, wu2025qwen, wei2025personalized, liu2025iidm, liu2025sad, dai2025latent, openai_gpt_image_1_5} models remain largely confined to pixel-level instruction alignment and the preservation of surface-level visual appearance, such as style, texture, or subjects, failing to capture the underlying abstract logic necessary for genuine metaphorical generation.
Transitioning from pixel-level reconstruction to metaphorical synthesis requires identifying deep-seated relational invariants across disparate domains and executing creative conceptual blending to induce emergent meaning.
Lacking an innate perception of such creative logic, current models cannot distill the metaphorical essence from a reference image and adapt it flexibly to novel contexts as humans do.


Research on visual metaphors has traditionally evolved along two parallel trajectories: interpretation and synthesis. 
Multimodal Large Language Models (MLLMs) have demonstrated fundamental cognitive abilities for metaphor interpretation, yet they struggle to parse non-literal semantics and deep symbolic relationships embedded in complex visual rhetoric without additional information or prompts~\cite{akula2023metaclue,kundu2025looking}. 
Simultaneously, synthesis methods remain predominantly text-driven, relying on mapping linguistic metaphors onto concrete objects through extensive textual prompts~\cite{chakrabarty2023spy, sun2025creative}. 
Despite their respective advancements, both paradigms converge on a shared limitation: an over-reliance on explicit, user-provided textual descriptions. 
This dependency creates a critical technical barrier to a more sophisticated creative capability—the ability for autonomously decoupling the underlying metaphorical logic from a visual reference and fluidly re-instantiating it within a novel context. 

To bridge this gap, we introduce the task of Visual Metaphor Transfer (VMT). 
Unlike conventional subject customization or style transfer that focuses on visual appearance, VMT necessitates deconstructing the ``creative essence'' from a reference image and re-materializing that abstract logic onto a user-specified target subject.
This paradigm shift presents two formidable challenges that push the boundaries of current generative AI: (1) Explicit metaphor modeling, which requires distilling domain-independent relational invariants from raw pixels into a structured representation; (2) Autonomous carrier adaptation, which demands the retrieval of a novel visual vehicle that not only aligns with the target subject’s attributes but also preserves the original cognitive tension to induce a fresh emergence of meaning. 
Addressing these challenges requires a transition from passive pixel synthesis to active, agentic visual reasoning.

To address these challenges, we propose a multi-agent framework for VMT, operationalizing Conceptual Blending Theory (CBT)~\cite{fauconnier1998conceptual, fauconnier2003conceptual} from cognitive linguistics into an executable computational paradigm. 
Central to our approach is the Schema Grammar ($\mathcal{SG}$), a novel structural representation that decouples abstract relational invariants from specific visual entities.
By encoding the intricate interplay between subjects, carriers, and semantic violations, $\mathcal{SG}$ provides a rigorous foundation for cross-domain logic re-instantiation.
Specifically, our framework executes VMT through a collaborative pipeline of specialized agents: (1) a \textit{perception agent} that distills the reference image into a structured schema; (2) a \textit{transfer agent} that maintains generic space invariance to autonomously discover contextually apt carriers for new subjects; and (3) a \textit{generation agent} that translates these logic blueprints into structured prompts for high-fidelity synthesis. 
Crucially, we introduce a hierarchical backtracing mechanism within a \textit{diagnostic agent}, which mimics a professional ``critic'' by identifying the root causes of failure across abstract logic, component selection, or prompt encoding. 
This closed-loop refinement ensures that the final output transcends mere pixel-level consistency to achieve profound logical alignment.
{We frame our approach primarily as a system contribution. Because end-to-end SOTA models lack the structured reasoning required for metaphor transfer, our core contribution lies in operationalizing Conceptual Blending Theory into a multi-agent, schema-driven workflow, thereby unlocking zero-shot creative capabilities that are fundamentally absent in base generative models.}
Our main contributions are summarized as follows:
\begin{itemize}[topsep=0em,leftmargin=1.5em,itemindent=0em,labelsep=1em]
\item {\textbf{Cognitive-Logic Formalization}: We operationalize conceptual blending theory (CBT) by proposing the ``Schema Grammar'' representation, providing a rigorous cognitive science foundation for cross-domain carrier matching and metaphor synthesis.}
\item {\textbf{Closed-Loop Multi-Agent Framework}: We develop a collaborative system encompassing perception, transfer, generation, and diagnostics. Notably, the proposed hierarchical backtracing mechanism significantly enhances generation reliability for complex metaphorical tasks.}
\item {\textbf{Superior Experimental Performance}: Our method outperforms existing baselines in terms of metaphor consistency, analogy appropriateness, and visual creativity.}
\end{itemize}

\section{Related Work}
\label{sec:relate}

\subsection{Visual Metaphor Understanding and Generation}
Visual metaphors serve as powerful rhetorical devices that convey abstract concepts through symbolic imagery. Unlike subject customization~\cite{ruiz2023dreambooth, galimage, avrahami2023break, xu2025b4m} or style transfer~\cite{zhang2022domain, hertz2024style, xu2026headrouter} approaches that focus on preserving visual appearance, metaphor understanding and generation requires capturing abstract symbolic relationships that convey meaning beyond literal visual similarity.
\revision{Classic work on analogy and structure-mapping~\cite{gentner1983structure} and its computational realization in the Structure-Mapping Engine (SME)~\cite{falkenhainer1989structure} establish that cross-domain transfer hinges on shared relational structure rather than surface attributes---a principle that motivates our schema-level Generic Space $G$ and informs subsequent CBT-based blending accounts.}
MetaCLUE~\cite{akula2023metaclue} shows that vision-language models struggle with metaphor understanding compared to literal images.
Building on this foundation, metaphor understanding works primarily focus on multimodal contexts, employing techniques including linking metaphor text to visual concepts with prompting~\cite{xu2024generating}, concept drift mechanisms~\cite{qian2025concept}, and prompt optimizer with reinforcement learning~\cite{fan2024prompt} to detect and interpret metaphorical content.
Complementing understanding approaches, recent works explore generating visual metaphors through text-driven synthesis. 
I-spy-a-metaphor~\cite{chakrabarty2023spy} propose a human–LLM–diffusion collaboration framework for generating visual metaphors from linguistic metaphors.
Creative Blends~\cite{sun2025creative} proposes an AI-assisted system that uses commonsense knowledge and LLMs to map abstract concepts to concrete objects and blend them via T2I models. 
TIAC~\cite{liao2024text} proposes a framework that maps abstract concepts to clear intents and semantic objects via LLMs to generate concept-aligned images.
Mind's eye~\cite{koushik2025mind} further explore self-evaluating visual metaphor generation framework with reinforcement learning.
{VisiBlends~\cite{chilton2019visiblends}, MetaMap~\cite{kang2021metamap}, and PopBlends~\cite{wang2023popblends} focus on supporting human conception through rule extraction or design-aided tools.}
However, these text-driven methods generate metaphors from linguistic input rather than learning from visual examples, requiring explicit textual specification of metaphorical concepts, which limits their ability to extract and transfer reusable metaphorical representations across different visual contexts.
In contrast, our method analyzes the core semantic meaning of metaphors and leverages the generic space from conceptual blending theory to transfer metaphorical representations to new subjects, achieving high-fidelity image-driven metaphor transfer.

\subsection{Image Generation with Multimodal LLMs}
Recent advances in multimodal large language models~\cite{li2022blip,seedream2025seedream, cao2025hunyuanimage, google_gemini_3_pro_image, openai_gpt_image_1_5} demonstrate remarkable capabilities in understanding and reasoning across vision and language modalities.
Building on these foundations, many image generation tasks leverage multimodal LLMs to decompose complex generation objectives into specialized subtasks and employ multi-agent frameworks to extend their functionality~\cite{shaham2024multimodal, sandoval2025editduet}.
For instance, SketchAgent~\cite{vinker2025sketchagent} utilizes LLMs with in-context learning to generate SVG strings that are subsequently rendered into sketches.
MCCD~\cite{li2025mccd} proposes a multi-agent scene parsing and hierarchical compositional diffusion framework to achieve image generation for complex multi-object prompts.
However, these methods primarily adopt a sequential execution paradigm where agents operate in a feed-forward manner without retrospective analysis.
In contrast, our method introduces a critic module that traces back to evaluate the output of each agent in previous steps and performs targeted refinement, enabling iterative improvement and achieving more faithful metaphor transfer that aligns with the intended symbolic meaning.
\section{Computational Modeling of Visual Metaphors}

To bridge human cognitive creativity and metaphorical transfer, we formalize the task by operationalizing conceptual blending theory (CBT). 
This section provides the theoretical foundation and structural representation for agentic reasoning.


\begin{figure}[t]
  \centering
  \includegraphics[width=0.75\linewidth]{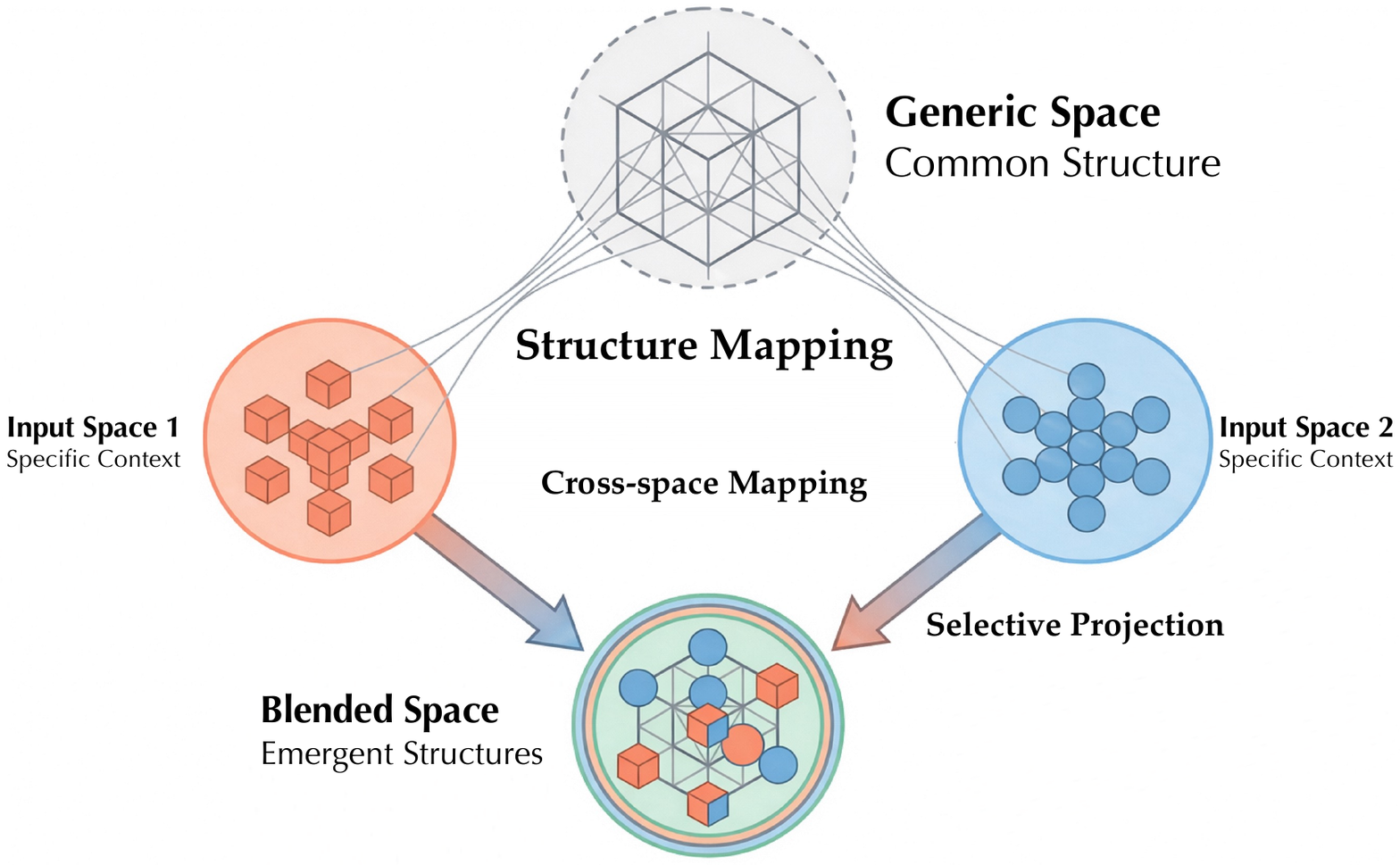}
  \vspace{-1.5em}
  \caption{{Illustration for conceptual blending theory.}}
  \label{fig:conceptual_blending}
    \vspace{-2.5em}
\end{figure}
\subsection{Conceptual Blending Spaces}
Conceptual blending theory, proposed by Fauconnier and Turner \shortcite{fauconnier1998conceptual, fauconnier2003conceptual}, posits that human creativity arises from integrating disparate mental spaces to generate novel meanings. 
A metaphor is not a simple linear mapping but a dynamic integration of four mental spaces, as illustrated in Fig.~\ref{fig:conceptual_blending}: 
(1) \textbf{Input Spaces} contain specific entities that provide raw content for the blend. 
Elements between these spaces are often linked by counterparts. 
(2) The \textbf{Generic Space} captures abstract, domain-independent relational invariants shared by both input spaces. 
It encodes the underlying logic (e.g., specific roles, frames, or schemata) that enables mapping between inputs. 
(3) The \textbf{Blended Space} is 
where elements from the inputs are selectively projected and integrated. 
Through composition, completion, and elaboration, this space gives rise to \textbf{emergent structures}--new relations and meanings absent from either input space alone.
\subsection{Structured Representation of Visual Metaphoric}
\begin{figure*}[!h]
  \centering
  \includegraphics[width=0.95\linewidth]{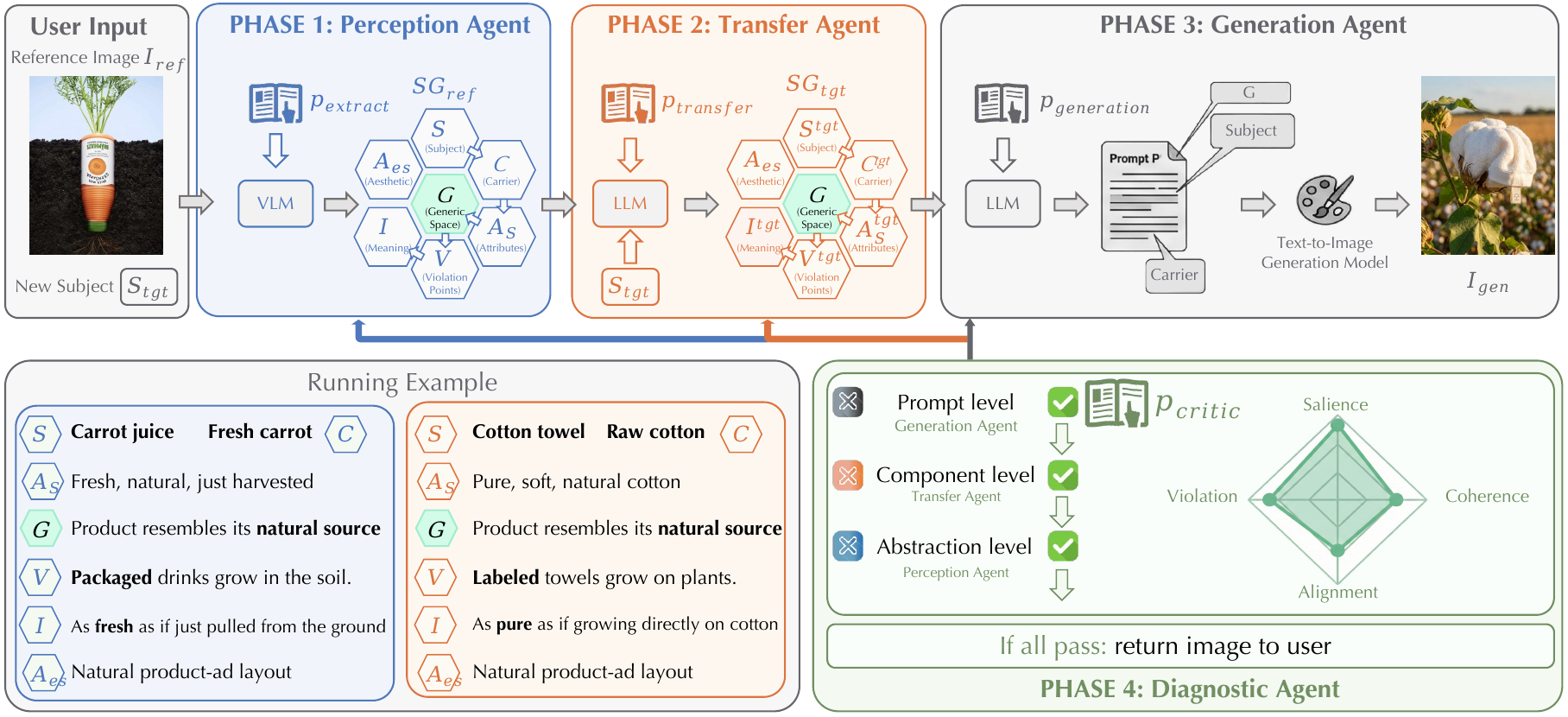}
  \vspace{-1.2em}
  \caption{\revision{Architecture of our Self-Reflective Agentic Framework for Visual Metaphor Transfer. The system consists of Perception, Transfer, Generation, and Diagnostic agents. 
  It transforms a given visual metaphor ($I_{ref}$) into a new target context ($I_{gen}$) by extracting structured schema representations and mapping them from the source to the target domain ($\mathcal{SG}_{ref} \rightarrow \mathcal{SG}_{tgt}$).
  In the second column, the quoted input text is the user-specified target subject $S_{tgt}$ (not a free-form metaphor description); the small explanatory text beneath is provided only to aid reader understanding and is \emph{not} available to the model during experiments.
  A hierarchical feedback loop ensures the generated output faithfully preserves the metaphor logic while adapting to the new subject matter.}}
  \vspace{-1.2em}
  \label{fig:pipeline}
\end{figure*}

\revision{We bridge psychological operations and computational reasoning by mapping these spaces into a structured \textbf{schema grammar} ($\mathcal{SG}$).
\rere{The 7-tuple is a direct operationalization of CBT~\cite{fauconnier1998conceptual, fauconnier2003conceptual}.}
We represent a visual metaphor as $\mathcal{SG}$ = $\{S, C, A_\sub{S}, \\A_\sub{es}, G, V, I\}$, where each element maps onto a blending component (\rere{see Fig.~\ref{fig:pipeline} for a running example of each element}):}
\begin{itemize}[topsep=0em,leftmargin=1em,itemindent=0em,labelsep=1em]
\item \revision{Entity instantiation ($\{S, C, A_\sub{S}, A_\sub{es}\}$): $S$ and $C$ instantiate the two Input Spaces---the subject (primary entity depicted) and the carrier (visual context or metaphorical vehicle). Visual metaphor embeds the subject into the carrier's domain. Inherent attributes $A_\sub{S}$ are the canonical properties of $S$ in its original domain and provide the baseline for detecting violations. Expressive attributes $A_\sub{es}$ capture image-level rhetorical atmosphere (e.g., lighting, mood, composition); preserving $A_\sub{es}$ ensures the transferred image retains the reference's expressive tone.}
\item \revision{Relational bridging ($\{G\}$): The Generic Space $G$ is the domain-independent relational invariant shared by $S$ and $C$ (functional, structural, relational, or emotional). Critically, $G$ is a \emph{constituent} of $\mathcal{SG}$, not a separate schema: $\mathcal{SG}_\sub{ref}$/$\mathcal{SG}_\sub{tgt}$ denote full schema grammars, while $G_\sub{ref}$/$G_\sub{tgt}$ denote only this relational bridging component.}
\item \revision{Synthesis operationalization ($\{V, I\}$): Selective projection and emergent structure in the Blended Space are realized through violation points $V$---semantic incongruities where $S$ transgresses norms of $C$, derived by contrasting $A_\sub{S}$ with $G$---and Emergent Meaning $I$, the high-level creative message induced by that cognitive tension.}
\end{itemize}
\revision{}{Removing any component breaks this CBT mapping: without $G$ there is no transferable invariant, without $V$ no cognitive tension, and without $I$ no target meaning. Our w/o CBT ablation (Sec.~\ref{sec:ablation}) confirms that this decomposition drives the observed gains.
By decoupling relational logic ($G, V, I$) from visual entities ($S, C$) while retaining $A_\sub{S}$ and $A_\sub{es}$, $\mathcal{SG}$ manipulates a metaphor's ``creative essence'' as a structured representation.}

\subsection{Task Formulation} 
\revision{Based on $\mathcal{SG}$, we formulate visual metaphor transfer as learning a mapping function $\mathcal{M}$ that migrates reference logic to a new subject. 
Given a reference schema $\mathcal{SG}_\sub{ref}$ and target subject $S_\sub{tgt}$, the framework must synthesize a target schema $\mathcal{SG}_\sub{tgt}$ such that:}
\begin{equation}
\begin{aligned}
\revision{\mathcal{M}(\mathcal{SG}_\sub{ref}, S_\sub{tgt})} & \revision{\rightarrow \mathcal{SG}_\sub{tgt},}  \\
s.t., {G}_\sub{tgt} & \equiv {G}_\sub{ref}.
\end{aligned}
\end{equation}
A transfer succeeds if \revision{$\mathcal{SG}_\sub{tgt}$ } preserves the reference's abstract relational logic ${G}$ while autonomously discovering a novel carrier $C_\sub{tgt}$ and violations $V_\sub{tgt}$ contextually appropriate for the new subject.
This formalization transforms VMT from pixel-level reconstruction into structured search and instantiation within the space of Schema Grammars.

\section{Method}
\label{sec:method}

In this section, we present our framework for visual metaphor transfer.
Our method decomposes metaphor transfer into four executable stages:
(1) The \textbf{Perception Agent} extracts an explicit reference schema \revision{$\mathcal{SG}_\sub{ref}$} from the input image.
(2) The \textbf{Transfer Agent} takes \revision{$\mathcal{SG}_\sub{ref}$} and the target subject $S_\sub{tgt}$, preserves the relational invariant $G$, and synthesizes \revision{$\mathcal{SG}_\sub{tgt}$} by selecting a new carrier and redesigning the violation points.
(3) The \textbf{Generation Agent} converts \revision{$\mathcal{SG}_\sub{tgt}$} into a detailed T2I prompt that grounds the subject, carrier, violation, emergent meaning, and aesthetic attributes.
(4) The \textbf{Diagnostic Agent} evaluates the generated image and performs prompt-, component-, or abstraction-level backtracking to correct failures in visual realization, carrier selection, or schema extraction.
The overall architecture is illustrated in Fig.~\ref{fig:pipeline}.
p


\subsection{Perception Agent}


The Perception Agent only analyzes the reference image, without adapting it to the target subject. Given $I_\sub{ref}$, it extracts a structured schema \revision{$\mathcal{SG}_\sub{ref}=\{S, C, A_\sub{S}, A_\sub{es}, G, V, I\}$}, making the implicit metaphor explicit. This stage separates surface entities from abstract relational logic, which direct prompting often fails to do.
We employ a Vision-Language Model (VLM) guided by chain-of-thought reasoning in the sequence:$S/C \rightarrow A_\sub{S} \rightarrow A_\sub{es} \rightarrow G \rightarrow V \rightarrow I$. The model first identifies the subject $S$ and carrier $C$, then extracts the subject attributes $A_\sub{S}$ and expressive attributes $A_\sub{es}$, such as layout, lighting, mood, and rhetorical atmosphere. It next isolates the Generic Space $G$, the domain-independent relation connecting $S$ and $C$. By contrasting $A_\sub{S}$ with the norms implied by $G$ and $C$, it derives the Violation Points $V$ that create cognitive tension, and finally infers the Emergent Meaning $I$. Formally, this extraction process is:
\begin{equation}\revision{\mathcal{SG}_\sub{ref} = \text{VLM}(I_\sub{ref}, p_\sub{extract}),}\end{equation}
where $p_\sub{extract}$ denotes the extraction-specific prompt. This decomposition turns an implicit creative concept into a manipulable reference schema. 
{The Perception Agent does not determine target adaptation; it only extracts reusable logic from the reference. By isolating $G$ while retaining $S$, $C$, $A_\sub{S}$, $A_\sub{es}$, $V$, and $I$, it provides the necessary input for the Transfer Agent to re-instantiate the same abstract logic with a new subject and carrier. A complete schema example is provided in the supplementary material.}


\subsection{Transfer Agent}
Given the reference schema \revision{$\mathcal{SG}_\sub{ref}$} and target subject $S_\sub{tgt}$, the Transfer Agent synthesizes a target schema \revision{$\mathcal{SG}_\sub{tgt}$} by preserving the abstract relational logic while re-grounding it in a new domain.
Unlike the Perception Agent, which analyzes only the reference image, this stage performs target-side schema synthesis: selecting a new carrier, redesigning violation points, and adapting the emergent meaning to the target subject.
This transfer is guided by VLM reasoning that keeps the Generic Space $G$ invariant across domains.

\textbf{a) Transfer Objective:}
\revision{The goal is to generate
$\mathcal{SG}_\sub{tgt} = \{S^\sub{tgt}, C^\sub{tgt},\\ A^\sub{tgt}_{S}, A^\sub{tgt}_\sub{es}, G, V^\sub{tgt}, I^\sub{tgt}\}$,
where $G$ is preserved from $\mathcal{SG}_\sub{ref}$, while $C^\sub{tgt}$, $A^\sub{tgt}_{S}$, $A^\sub{tgt}_\sub{es}$, $V^\sub{tgt}$, and $I^\sub{tgt}$ are re-instantiated for the target domain.}
This retains the reference metaphor's relational logic while realizing it through a target-specific visual configuration.

\textbf{b) Reasoning process:}
We prompt the VLM to perform relational transfer through:
$G \rightarrow A^\sub{tgt}_{S} \rightarrow C^\sub{tgt} \rightarrow V^\sub{tgt} \rightarrow I^\sub{tgt}$.
It first analyzes the domain-independent nature of $G$, then profiles the inherent attributes and functional or symbolic roles of $S_\sub{tgt}$.
Next, it searches for a cross-domain carrier $C^\sub{tgt}$ sharing the same Generic Space $G$, designs violation points $V^\sub{tgt}$ that mirror the reference violation logic, and aligns $I^\sub{tgt}$ with the target context.

This transfer process is formalized as:
\begin{equation}
\revision{\mathcal{SG}_\sub{tgt} = \text{VLM}(\mathcal{SG}_\sub{ref}, S_\sub{tgt}, p_{transfer}),}
\end{equation}
where $p_{transfer}$ specifies the relational reasoning chain and invariance constraint on $G$.

The resulting \revision{$\mathcal{SG}_\sub{tgt}$} serves as the conceptual blueprint for visual synthesis.
By keeping $G$ invariant while re-instantiating $C^\sub{tgt}$, $V^\sub{tgt}$, and $I^\sub{tgt}$, the Transfer Agent enables target-specific metaphor construction instead of directly copying the reference schema.
This distinction is validated by the \textit{w/o Transfer Agent} ablation, where directly feeding \revision{$\mathcal{SG}_\sub{ref}$} to generation often preserves reference fragments but fails to form a coherent target-domain metaphor.

\subsection{Generation Agent}

\revision{Given the target schema $\mathcal{SG}_\sub{tgt}$, the Generation Agent converts its structured components into a T2I prompt for visual realization.}
Unlike the Transfer Agent, which performs schema-level reasoning, this stage translates the subject, carrier, violation, emergent meaning, and expressive attributes into executable generative instructions:
\begin{equation}
\revision{P = \text{LLM}(\mathcal{SG}_\sub{tgt}, p_\sub{generation}),}
\end{equation}
where $p_\sub{generation}$ denotes the generation-specific system prompt.

The prompt construction emphasizes $C_\sub{tgt}$ for scene layout, $V_\sub{tgt}$ for conceptual tension, and $I_\sub{tgt}$ with $A^\sub{tgt}_\sub{es}$ for style and atmosphere.
Finally, the target metaphor image is synthesized by a pre-trained generator, which means  
$
I_\sub{gen} = \text{Gen}(P).
$

\subsection{Diagnostic Agent}
\label{sec:feedback}
The generated image $I_\sub{gen}$ may still contain visual or conceptual errors caused by prompt ambiguity, unsuitable component choices, or incorrect schema abstraction.
To address this, we introduce a VLM-based Diagnostic Agent that evaluates the result and performs hierarchical refinement.
We further extend the Diagnostic Agent with a safety and alignment check: the LLM evaluates the outputs against safety guidelines and rejects or revises schemas that may induce malicious, offensive, or harmful alignments.
Unlike the Generation Agent, which only realizes a fixed \revision{
$\mathcal{SG}_\sub{tgt}$}, this stage decides whether errors should be corrected at the prompt, component, or abstraction level.

\textbf{a) Diagnostic dimensions.}
The VLM examines $I_\sub{gen}$ across four complementary dimensions:\\[-13pt]
\begin{itemize}[leftmargin=20pt]
    \item \textbf{Subject Salience:} whether $S^\sub{tgt}$ is recognizable and preserves its core attributes;
    \item \textbf{Violation Realization:} whether $V^\sub{tgt}$ is visually explicit and structurally coherent;
    \item \textbf{Relational Coherence:} whether the Generic Space $G$ is successfully instantiated and perceptible; and
    \item \textbf{Meaning Alignment:} whether the image conveys the intended $I^\sub{tgt}$ without negative ambiguity{, malicious implication, or offensive association.}
\end{itemize}
Rather than producing numerical scores, the VLM outputs qualitative issue descriptions and assigns each failure to a backtracking level.
{We provide the diagnostic decision table in the supplementary materials, mapping failure types to their corresponding revision levels and strategies.}

\textbf{b) Hierarchical backtracking and refinement.}
{Based on the diagnostic output, we perform cascaded attribution through three levels governed by strict conditional failure triggers.}
\textit{Prompt-level} backtracking is triggered when the VLM determines the conceptual structure \revision{($\mathcal{SG}_\sub{tgt}$)} is sound, but visual details fail (e.g., $P$ under-specifies spatial relations or atmosphere);
we revise $P$ and regenerate.
\textit{Component-level} {backtracking is triggered when the VLM detects unbridgeable domain gaps that make the outputs unrecognizable or visually ambiguous, indicating that $C^\sub{tgt}$ or $V^\sub{tgt}$ is unsuitable;} we return to the Transfer Agent to replace the visual carrier or redesign the violation.
\textit{Abstraction-level} {backtracking is triggered when component modifications repeatedly fail (exceeding a local iteration threshold), suggesting that the underlying metaphorical logic ($G$)} was extracted at an improper abstraction level; we revisit \revision{$\mathcal{SG}_\sub{ref}$} and re-extract the relational logic.
This refinement is formulated as:
\begin{equation}
\revision{I_\sub{final} = \text{Refine}(I_\sub{gen}, \mathcal{SG}_\sub{tgt}, p_\sub{critic}; \tau),}
\end{equation}
where $p_\sub{critic}$ denotes the diagnostic prompt and $\tau$ the iteration threshold.
The loop stops when the diagnostic feedback indicates satisfactory quality or reaches the maximum iteration limit, yielding $I_\sub{final}$.
\section{Experiments}
\label{sec:experiments}

\begin{figure*}[!h]
  \centering
  \includegraphics[width=1\linewidth]{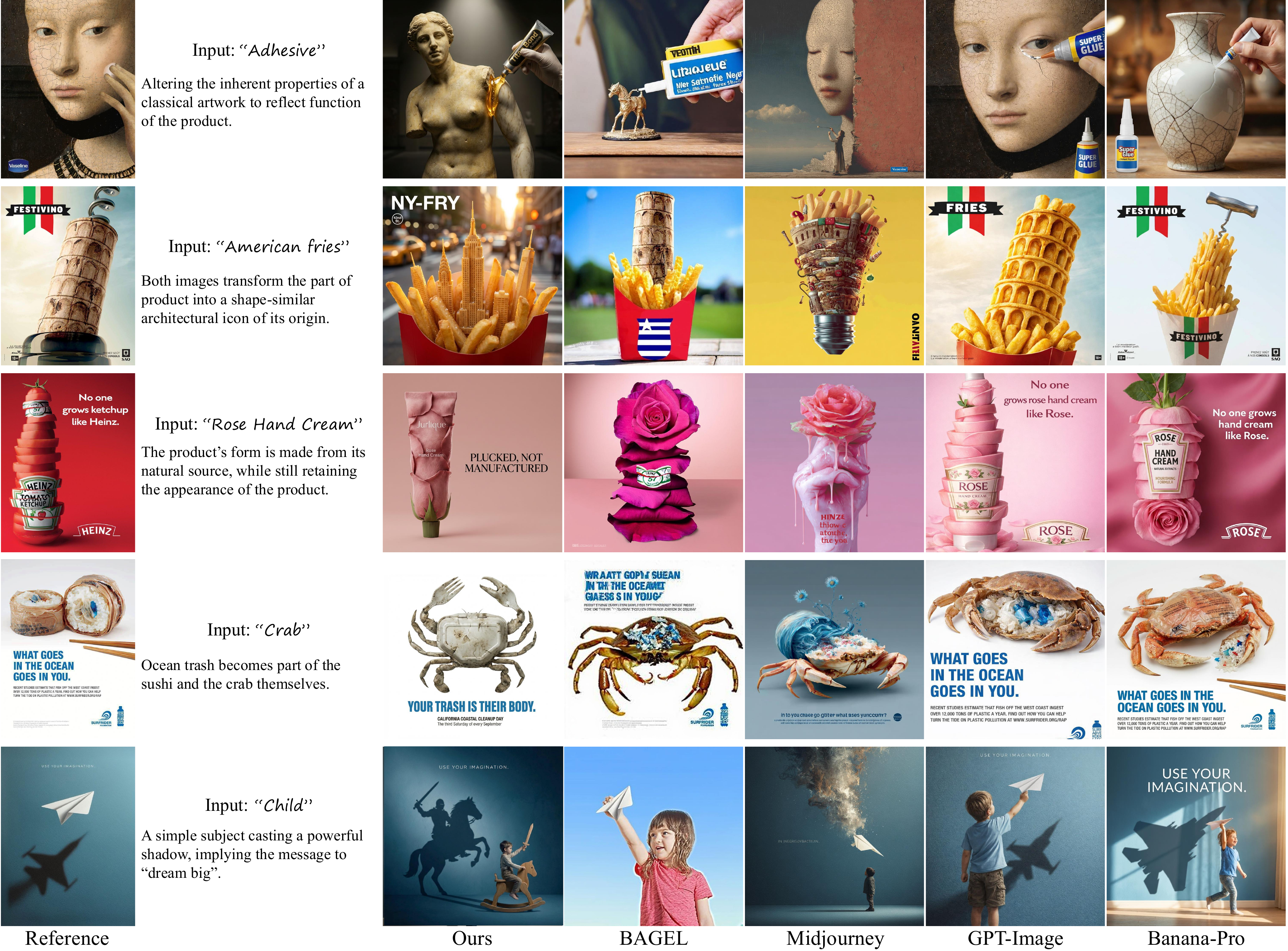}
  \vspace{-2.5em}
  \caption{Qualitative comparison with baseline methods.}
  \vspace{-0.4em}
  \label{fig:main_compare}
\end{figure*}
\begin{figure*}[!htbp]
  \centering
  \includegraphics[width=\linewidth]{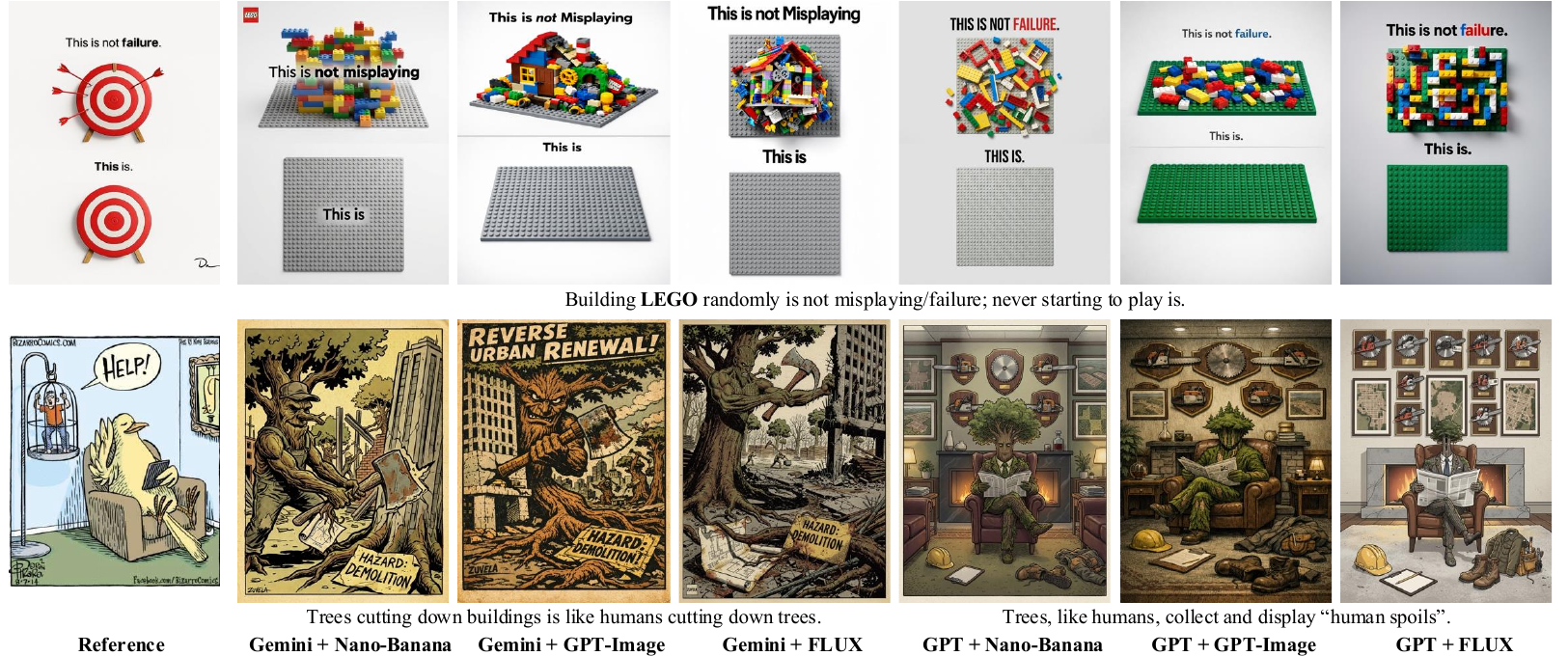} 
  \caption{\revision{\textbf{Qualitative comparison of different backbone combinations.} 
  We validate the framework's generalizability by pairing different LLMs (Gemini, GPT) with various T2I models (Banana-pro, GPT-Image, FLUX). FLUX remains essentially competitive with commercial generators.}}
  \vspace{-1.2em}
  \label{fig:ablation_gpt}
\end{figure*}

We first introduce the experiment settings in Sec.~\ref{sec:setting}, and the present qualitative, quantitative, and human evaluation results in Secs.~\ref{sec:Qualitative}, \ref{sec:Quantitative} and \ref{sec:human}. Finally, we conduct an ablation study and generalizability analysis in Secs.~\ref{sec:ablation} and~\ref{sec:Generalizability}.

\subsection{Experimental settings}
\label{sec:setting}
\textit{Baselines.}
We compare our approach against state-of-the-art multimodal image generation models with integrated visual understanding and reasoning capabilities, as metaphor transfer inherently requires analyzing the source image's creative concept before generating the target. \revision{We evaluate image generation baselines \textbf{BAGEL}~\cite{deng2025emerging}, \textbf{Midjourney}~\cite{midjourney}, \textbf{GPT-Image}~\cite{openai_gpt_image_1_5}, and \textbf{Banana-pro}~\cite{google_gemini_3_pro_image}, and use these names consistently throughout the paper.}
\revision{We further compare against I-spy-a-metaphor~\cite{chakrabarty2023spy}. Because that method generates visual metaphors from linguistic metaphors, we prepend a linguistic-metaphor extraction stage over the reference image to align it with our VMT pipeline, then evaluate under the same metrics.}

\textit{Datasets.} 
We curated a diverse dataset of 126 visual metaphors from the internet, including product ads (32), memes (33), film posters (15), comics (10), and other creative works (36). This heterogeneous collection spans multiple domains to comprehensively test our framework’s generalization across various metaphorical styles and compositions.

\textit{Metrics.}
Unlike conventional image evaluation methods such as CLIP~\cite{ilharco_gabriel_2021_5143773} or DINO~\cite{liu2025groundingdino} that primarily assess low-level visual features or semantic similarity, metaphor transfer requires evaluating high-level conceptual reasoning and abstract creative alignment, which necessitates the use of VLMs capable of understanding complex analogical relationships.
We employ three frontier VLMs, Gemini-3-pro, GPT-5.2, and Claude-Sonnet-4.5 to assess generated images across multiple dimensions using 10-point scales : (1) Metaphor Consistency (MC), which measures whether the target metaphor preserves the core metaphor logic of the source; (2) Analogy Appropriateness (AA), which evaluates the validity of functional and formal correspondences between the carrier and target subject; and (3) Conceptual Integration (CI), which assesses whether the fusion between the subject and carrier appears natural and harmonious. We also evaluate image aesthetic quality with a SigLip-based predictor~\cite{zhai2023sigmoid} to ensure visual appeal.
Note that these evaluation VLMs are distinct from the VLM used in our iterative refinement process (Section~\ref{sec:feedback}), ensuring independent assessment of generation quality.
\revision{Using three judges from different providers mitigates single-model bias; we additionally report Qwen-VL judgments in the supplementary material, where our method remains strongest. Per-metric human--VLM agreement is likewise reported in the supplementary material.}
The complete VLM evaluation prompts and the validation of VLM-as-judge reliability are provided in the supplementary material.

\textit{Implementation.}
{We employ Gemini-3-pro as both the VLM and LLM in our pipeline, and utilize Banana-pro for image generation.} The iteration threshold $\tau$ is set to 5 to balance refinement quality and computational efficiency. $p_\text{extract}$, $p_\text{transfer}$, $p_\text{generation}$ and $p_\text{critic}$ are provided in the supplementary material.
\revision{On the 126-metaphor set, the average number of calls per agent is Perception 1.14, Transfer 1.25, Generation 2.19, and Diagnostic 2.19 (ideal value 1); sequential dependence causes call counts to increase along the pipeline. Average wall-clock time is 4\,min\,11\,s with about 8.3k tokens per sample (see supplementary Consumption). To assess reproducibility beyond commercial endpoints, we also re-run the full pipeline with open-source, version-pinned models---Qwen (LLM) and Qwen-VL (VLM) for Perception/Transfer/Diagnostic, and FLUX for generation---and report Strong Prompt / w/o CBT under the same open-source stack (supplementary Table); although absolute scores drop slightly, relative gains of our method persist, confirming that improvements stem from the framework rather than a black-box model.}

\begin{table}[t]
\centering
\caption{Quantitative evaluation results. Best results in \textbf{bold}.}
\vspace{-1em}
\label{tab:main_compare}
\scalebox{0.7}{
\begin{tabular}{ccccccccccc}
\toprule
\multirow{2}{*}{\textbf{Methods}} & \multicolumn{3}{c}{\textbf{Gemini-3-pro}} & \multicolumn{3}{c}{\textbf{GPT-5.2}} & \multicolumn{3}{c}{\textbf{Claude-4.5}} & \multirow{2}{*}{Aes.$\uparrow$}\\
\cline{2-10}
& MC$\uparrow$ & AA$\uparrow$ & CI$\uparrow$ & MC & AA & CI & MC & AA & CI \\ \hline
BAGEL       &5.17 &4.55 &5.05  &6.21 &5.83 &6.07  &6.05 &5.58 &5.95 &4.77 \\
Midjourney  &5.33 &5.57 &6.09  &6.33 &6.46 &6.24  &6.51 &5.94 &6.06 &5.22 \\
GPT-Image   &8.08 &7.59 &7.47  &7.71 &7.65 &7.54  &7.95 &7.39 &7.51 &5.63 \\ 
Banana-pro  &8.75 &7.68 &7.33  &7.95 &7.77 &7.37  &8.08 &7.42 &7.74 &5.57 \\
\revision{I-spy-a-metaphor} &\revision{8.86} &\revision{8.02} &\revision{7.49}  &\revision{8.14} &\revision{7.98} &\revision{7.55}  &\revision{8.52} &\revision{7.83} &\revision{7.94} &\revision{5.48} \\
\hline
Direct Prompt &8.79 &8.03 &7.63  &8.13 &7.96 &7.69  &8.44 &7.85 &7.92 &5.59 \\
 {Strong Prompt}  & {9.02} & {8.13} & {7.68}  & {8.38} & {8.04} & {7.72}  & {8.59} & {7.95} & {8.02} & {5.61} \\
w/o CBT  &8.91 &8.09 &7.58  &8.33 &8.01 &7.71  &8.56 &7.89 &7.97 &5.61 \\
 {w/o Trans.}  & {8.97} & {8.11} & {7.47}  & {8.36} & {8.07} & {7.41}  & {8.58} & {7.97} & {7.49} & {5.44} \\ 
w/o Diag.  &9.14 &8.47 &8.33  &8.44 &8.33 &8.29  &8.62 &8.38 &8.19 &5.63 \\ \hline
Ours       &\textbf{9.31} &\textbf{8.97} &\textbf{8.76}  &\textbf{8.62} &\textbf{8.51} &\textbf{8.58}  &\textbf{8.73} &\textbf{8.61} &\textbf{8.36} &\textbf{5.68} \\ 
\bottomrule
\end{tabular}
}
\vspace{-2em}
\end{table}


\subsection{Qualitative comparisons}
\label{sec:Qualitative}
As shown in Fig.~\ref{fig:main_compare}, our method excels at decoupling abstract creative logic from source domains and re-materializing it within novel targets. While SOTA models like GPT-Image and Banana-pro are visually proficient, they rely on surface-level manipulation rather than decoding underlying metaphoric logic.
For instance, in the ``American Fries'' task, these baselines merely substitute components without grasping the ``regional architectural landmark with similar shape'' metaphor, whereas in the ``Rose Hand Cream'' case, they erroneously preserve the ``sliced'' geometry from the reference, which is semantically incongruent with the new subject.
This conceptual deficiency extends to structural integration: in the ``Crab'' example, baselines scatter trash as background clutter instead of merging it into the organism's anatomy, and in the ``Child'' scene, they fail to project a ``powerful shadow'' to convey the intended ``dream big'' message. Furthermore, models like BAGEL and Midjourney tend to generate results from scratch, leading to a loss of metaphoric alignment.
In contrast, our method achieves superior semantic reasoning by successfully synthesizing New York landmarks that mirror the geometric form of fries while signifying their origin, blending organic rose textures with traditional packaging, and seamlessly embedding plastic waste into the crab's biological structure.
By accurately mapping abstract relationships, our approach demonstrates a unique capacity to re-materialize abstract creative intents while ensuring high-level conceptual consistency.

\subsection{Quantitative comparisons}
\label{sec:Quantitative}
As Tab.~\ref{tab:main_compare} shows, our method consistently outperforms all baselines across three frontier VLMs and the aesthetic predictor. Notably, we achieve the most significant improvement in the AA metric (a 16.8\% increase over the runner-up) demonstrating that our proposed Metaphor Transfer Agent effectively identifies metaphorically consistent visual carriers that best match new subjects. Beyond AA, our approach maintains superior scores in MC and CI, while also securing the highest aesthetic score (5.68). This consensus among evaluators (Gemini, GPT, and Claude) underscores our framework’s robustness in generating logically sound and visually harmonious metaphorical images without sacrificing artistic quality.
\revision{Relative to the adapted I-spy-a-metaphor, our method improves MC/AA/CI under all three judges, indicating that schema-driven transfer and closed-loop diagnosis outperform linguistic-metaphor co-creation when aligned to the VMT setting.}

\subsection{Human evaluation study}
\label{sec:human}
{To validate the perceptual quality and creative effectiveness between our method and the baselines, we conducted a comprehensive human evaluation study with 100 participants (age 14-55, 45 male, 55 female)}. The study comprises two tasks. 

\begin{figure}[!htbp]
  \centering
  \includegraphics[width=\linewidth]{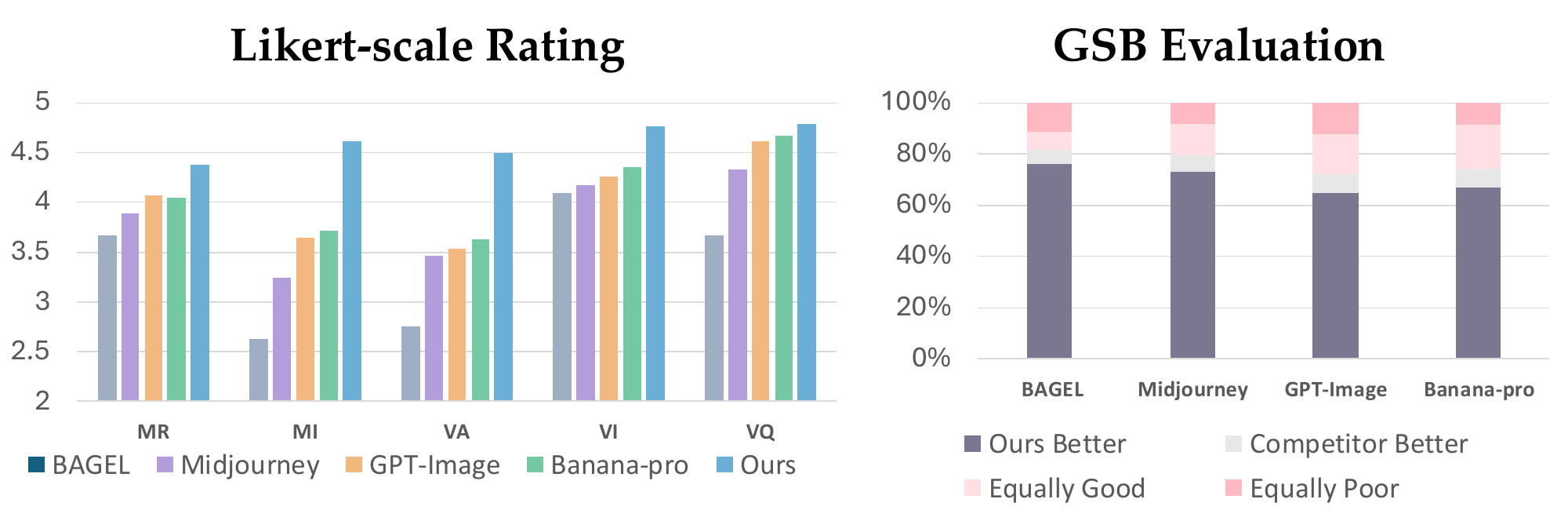}
 \vspace{-2.2em}
  \caption{{Human evaluation study.}}
 \vspace{-1em} 
  \label{fig:user_study_all}
\end{figure}

In Task 1, each participant evaluates {45} images generated by each method (ours and 4 baselines, totaling {225} images per participant) independently along five dimensions using 5-point Likert scales: (1) Metaphor Recognizability (MR); (2) Metaphor Ingenuity (MI); (3) Violation Appropriateness (VA); (4) Visual Integration (VI); and (5) Overall Visual Quality (VQ). The detail of the definitions are in the supplementary material. 
As shown in the left of Fig.~\ref{fig:user_study_all}, our method consistently outperforms all baselines across all five dimensions. Notably, it achieves a significant lead in MI ({4.61}) and VA ({4.49}), indicating that our framework produces more creative and purposeful metaphorical designs than SOTA models like Banana-pro and GPT-Image. Furthermore, our approach attains the highest scores in VI ({4.77}) and VQ ({4.79}), confirming that our focus on conceptual reasoning does not compromise aesthetic fidelity. These results collectively demonstrate our method's superior capability in synthesizing metaphorical images.

In Task 2, we conduct GSB (Good/Same/Bad) evaluation to assess user preference by asking participants which image in each pair delivered a more compelling metaphorical message. As shown in the right of Fig.~\ref{fig:user_study_all}, our method is consistently favored by participants, securing over 60\% ``Ours Better'' ratings across all baseline comparisons.
Notably, our approach outperforms strong commercial competitors such as GPT-Image and Banana-pro in {64.44\%} and {67.11\%} of cases, respectively, while being judged as inferior in fewer than 10\% of pairs. This preference margin widens further against Midjourney ({72.89\%}) and BAGEL ({76.22\%}), demonstrating our framework's metaphorical syntheses are significantly more resonant and conceptually effective than existing SOTA models.



\begin{figure}[t]
  \centering
  \includegraphics[width=\linewidth]{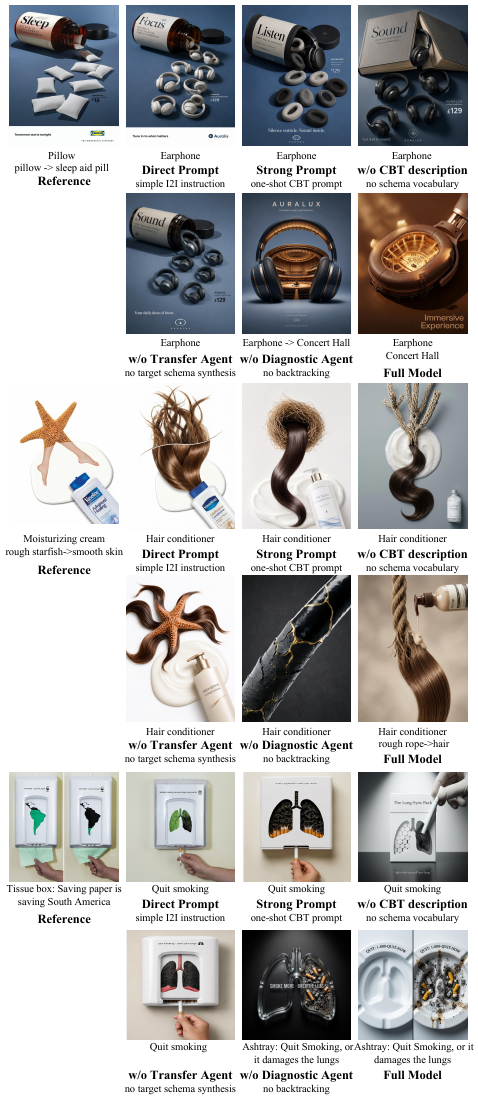} 
 \vspace{-1.2em}
  \caption{{\textbf{Qualitative comparison of ablation variants.}}
}
  \label{fig:qualitative_ablation}
\end{figure}

\subsection{Ablation study}
\label{sec:ablation}
{We conduct qualitative and quantitative comparisons across ablation variants, as shown in Fig.~\ref{fig:qualitative_ablation} and Tab.~\ref{tab:main_compare}. 
To disentangle prompt strength, CBT-based schema description, staged transfer, and diagnostic refinement, all variants use the same reference image, target subject, generator, and evaluation protocol, differing only in the enabled components or provided descriptions.}

{{\textit{Direct prompt}.}}
{We first evaluate a direct I2I prompting baseline using Banana-pro~\cite{google_gemini_3_pro_image}. 
The model receives only a simple instruction: ``\textit{Understand this advertisement image, analyze its metaphors and creative ideas, and transfer this creative idea to the new product}.''} 
Without structured guidance, the model often falls back to literal object substitution or salient-shape copying, while missing the underlying relational mechanism, as shown in Tab.~\ref{tab:main_compare}.
The drops in MC, AA, and CI indicate limited metaphor understanding under simple prompting.

{{\textit{Strong prompt with full CBT description}.}}
{To test whether the gains mainly come from stronger prompt engineering, we introduce a manually designed strong-prompt baseline with a full CBT-style task description. 
The prompt explicitly instructs the LLM to analyze the reference image in terms of subject $S$, carrier $C$, relational invariant $R$, violation point $V$, and emergent meaning $M$, and to directly convert this one-shot analysis into a text-to-image prompt.}
{Although the stronger prompt improves metaphor awareness compared to direct prompting, it remains a single-stage approach and therefore often yields unstable carrier selections or incomplete relational transfer. As illustrated in Tab.~\ref{tab:main_compare}, it may capture partial visual or functional cues, yet still fails to consistently preserve the cross-domain metaphorical structure. These results indicate that prompt strength alone does not fully account for the observed improvements.
}

{{\textit{W/o CBT description} .}}
{We then remove the CBT-based schema description while keeping the overall multi-stage prompting process (including iteration). 
Without explicit notions of subject, carrier, relational invariant, violation, and emergent meaning, the model lacks a structured vocabulary for decomposing and transferring the reference metaphor.}
{As shown in Tab.~\ref{tab:main_compare}, this variant can generate visually plausible results, but the outputs tend to rely on generic scene composition, superficial texture transfer, or loosely related causal cues rather than preserving the intended metaphorical logic. 
The performance drop relative to the full method shows that merely introducing iteration without the CBT schema is inferior, and that the CBT description is important for stabilizing metaphor decomposition and transfer.}

{{\textit{W/o Transfer Agent} .}}
{We next remove the Transfer Agent while retaining schema extraction. 
The Perception Agent extracts the reference schema $\mathcal{SG}_\sub{ref}$, which is directly fed to an LLM to produce the T2I prompt, without synthesizing a target schema $\mathcal{SG}_\sub{tgt}$. 
Thus, this variant has access to the reference schema but lacks explicit target-side carrier selection, violation redesign, and relational re-instantiation.}
{Although it preserves more reference information than prompt-only variants, it adapts the schema poorly to the target domain, leading to unsuitable carriers, superficial attribute transfer, or incomplete instantiation of the reference mechanism, as shown in Tab.~\ref{tab:main_compare}. 
The AA drop indicates that target-side schema synthesis is necessary for constructing coherent transferred metaphors.}

{\textit{W/o Diagnostic Agent}.}
{This variant keeps the Perception, Transfer, and Generation Agents but removes Phase~4, generating once from $\mathcal{SG}_\sub{tgt}$ without prompt-, component-, or abstraction-level revision.}
Without hierarchical diagnosis, semantic and structural errors persist, including misleading source-context associati-

\noindent ons, ambiguous carrier realization, and failures to enforce the required visual layout, as shown in Tab.~\ref{tab:main_compare}.
The consistent score drops confirm the importance of closed-loop refinement.

\textit{{Full model.}}
{The full model combines all components: Perception extracts $\mathcal{SG}_\sub{ref}$, Transfer synthesizes $\mathcal{SG}_\sub{tgt}$ by preserving the relational invariant while re-instantiating the carrier and violation, Generation converts $\mathcal{SG}_\sub{tgt}$ into a T2I prompt, and Diagnosis performs hierarchical backtracking when needed.}
Across Rows~1--5 in Tab.~\ref{tab:main_compare}, the full model better preserves the intended relational logic, selects suitable target-domain carriers, and enforces the required visual structures, demonstrating the necessity of each component for high-quality metaphor transfer.

\revision{\paragraph{User study on ablations}
To verify that these component-wise gains are also perceptible to human viewers, we extend the human evaluation protocol to all ablation variants, participants rate the outputs of each ablation variant along the same five dimensions on 5-point Likert scales. As shown in Tab.~\ref{tab:ablation_user_study}, our full model receives the highest ratings on all dimensions. Direct Prompt is rated lowest, especially on MI (2.75) and VA (2.96), while Strong Prompt and w/o CBT plateau around 3.7 on these conceptual dimensions, confirming that neither stronger prompting nor iteration without the CBT schema suffices. Even against the strongest variant (w/o Diag.), the remaining gaps concentrate on MI (+0.51) and VA (+0.53), indicating that the perceived advantage stems from schema-driven reasoning and closed-loop diagnosis rather than rendering quality. }

\begin{table}[t]
\centering
\caption{User study results for ablations. Best results in \textbf{bold}.}
\vspace{-1em}
\label{tab:ablation_user_study}
\scalebox{1}{
\begin{tabular}{cccccc}
\toprule
\textbf{Methods} & MR$\uparrow$ & MI$\uparrow$ & VA$\uparrow$ & VI$\uparrow$ & VQ$\uparrow$ \\
\hline
Direct Prompt  & 3.62 & 2.75 & 2.96 & 3.75 & 3.65\\
Strong Prompt  & 3.97 & 3.69 & 3.77 & 3.89 & 4.09\\
w/o CBT        & 3.88 & 3.66 & 3.68 & 4.04 & 4.12\\
w/o Trans.     & 4.01 & 3.71 & 3.83 & 4.33 & 4.25\\
w/o Diag.      & 4.26 & 4.05 & 3.98 & 4.61 & 4.48\\
\hline
Ours           & \textbf{4.47} & \textbf{4.56} & \textbf{4.51} & \textbf{4.74}  & \textbf{4.76} \\
\bottomrule
\end{tabular}
}
\vspace{-2em}
\end{table}

\subsection{Generalizability analysis}
\label{sec:Generalizability}
To verify the robustness and model-agnostic nature of our framework, we evaluate it across different LLM and T2I generator combinations. As shown in Fig.~\ref{fig:ablation_gpt}, we test two reasoning backbones (Gemini, GPT-4) paired with three rendering engines (Banana-pro, GPT-Image, FLUX).
\revision{In particular, FLUX results remain essentially competitive with commercial generators, supporting that the pipeline is not tied to a single proprietary image model.}

\textit{Consistency in metaphorical mapping.} 
The top row shows the ``LEGO'' scenario, where the core metaphor contrasts ``disorganized effort'' with ``total inaction''. Regardless of the T2I model, the framework translates the reference's ``missed target'' concept into a ``chaotic LEGO pile'', and ``total inaction'' into an ``empty baseplate''. This consistency shows that our method preserves the metaphor's logical structure across visual generators.

\textit{Diversity in narrative interpretation.} 
The bottom row (``Protecting Forests'') shows how different LLMs shape the creative narrative while maintaining visual coherence. 
Gemini-driven variants interpret role reversal as \textit{active retaliation}, prompting anthropomorphic trees to demolish urban buildings with axes.
GPT-driven variants interpret it as \textit{cultural satire}, depicting civilized trees displaying chainsaws as ``hunting trophies.''
Despite these divergences, all T2I models faithfully render their prompts. This demonstrates that our framework supports creative flexibility in reasoning while preserving high-fidelity visual execution during generation.

\begin{figure}[t]
  \centering
  \includegraphics[width=\linewidth]{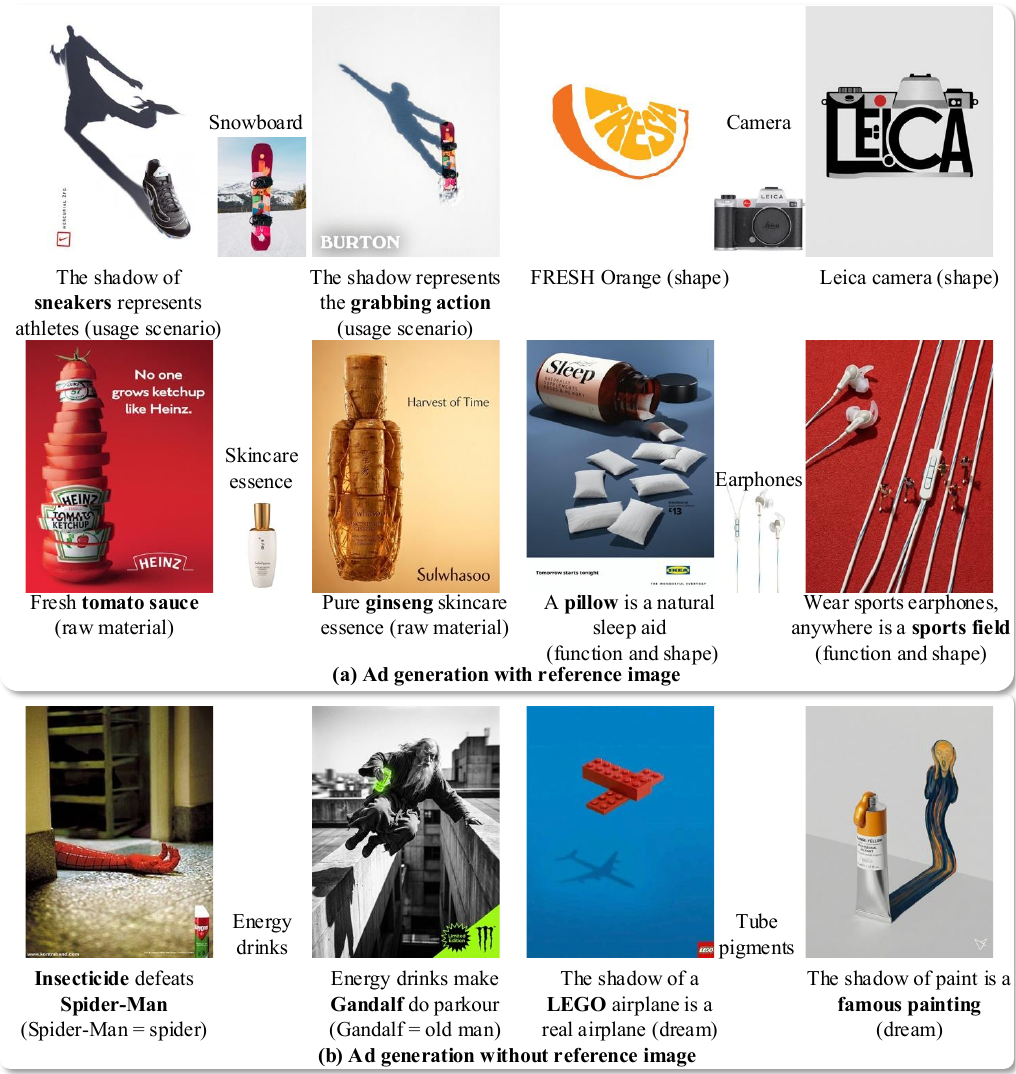} 
 \vspace{-1.2em}
  \caption{\textbf{Application of product advertisements for multiple-reference-guided scenarios.} Our method can flexibly handle both visual-to-visual and text-to-visual creative workflows.}
  \label{fig:application_ad}
\end{figure}

\begin{figure}[!htbp]
  \centering
  \includegraphics[width=0.95\linewidth]{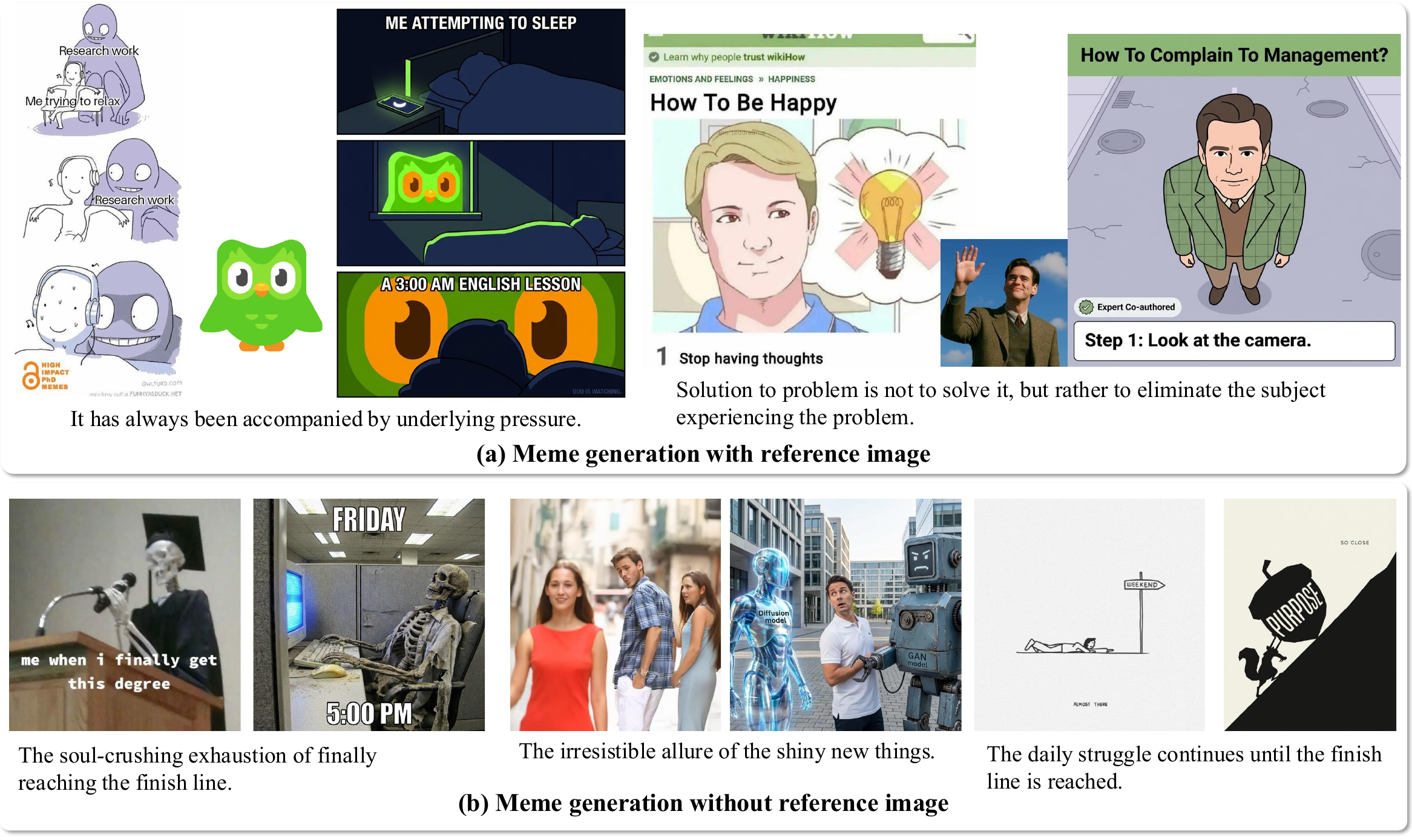}
  \vspace{-1.em}
  \caption{\textbf{Application of meme image generation.} }
  \label{fig:application_meme}
\end{figure}

\begin{figure}[!htbp]
  \centering
  \includegraphics[width=0.95\linewidth]{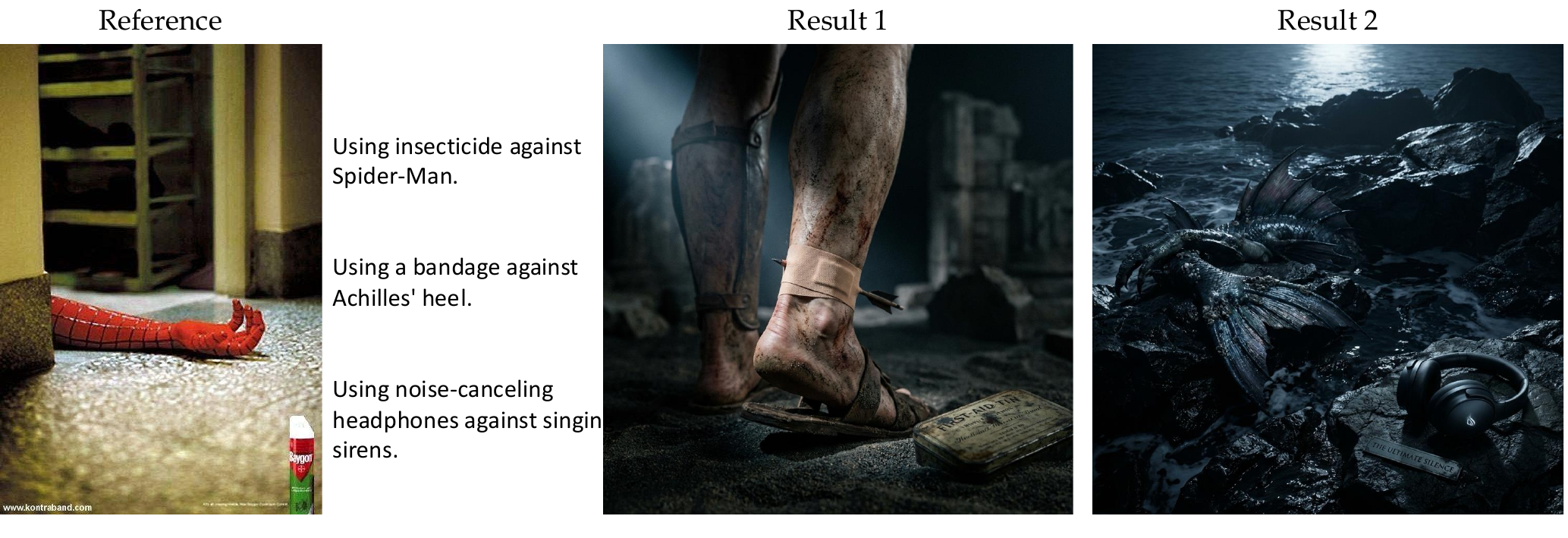}
  \caption{Badcase. Cognitive overload and obscure symbolism.}
  \label{fig:badcase}
\end{figure}

We presented more applications for product advertisements and meme image generation in Figs.~\ref{fig:application_ad} and \ref{fig:application_meme}, bad cases in Fig.~\ref{fig:badcase}. For corresponding discussions, please refer to the supplementary materials.




\section{Conclusions}
We introduced visual metaphor transfer, a task that goes beyond pixel-level editing by extracting the underlying metaphor logic from a reference image and re-instantiating it on a user-specified new subject. To achieve this, we formalize metaphor structure with Schema Grammar \revision{$\mathcal{SG}$} and construct a closed-loop multi-agent pipeline comprising: a Perception Agent that extracts the schema, a Transfer Agent that preserves the Generic Space $G$ while finding a new carrier, a Generation Agent that turns the schema into prompts, and a Diagnostic Agent that backtraces failures across prompt, component, and abstraction levels.
Experiments show that this design improves metaphor consistency, analogy appropriateness, and conceptual integration compared with baselines.

\begin{acks}
We thank the reviewers for their insightful comments and suggestions.
This work was partly supported by the National Natural Science Foundation of China under No. 62572458, National Science and Technology
Council, Taiwan under Grant 113-2221-E006-161-MY3, and the Deutsche Forschungsgemeinschaft (DFG, German Research Foundation) under Germany's Excellence Strategy–EXC 2117–422037984.
\end{acks}

\clearpage
\bibliographystyle{ACM-Reference-Format}
\bibliography{sample-base}

\end{document}


\title{Supplementary Materials for ``Beyond Pixels: Visual Metaphor Transfer via Schema-Driven Agentic Reasoning''}

\maketitle

\tableofcontents

\section{Pseudocode of Our Method}

We summarize the complete execution pipeline of our Visual Metaphor Transfer framework in Algorithm~1. The process initiates with the Perception Agent, which distills the reference image ($I_\sub{ref}$) into a structured Schema Grammar \revision{($\mathcal{SG}_\sub{ref}$)} by isolating the domain-independent Generic Space $G$. Subsequently, the Transfer Agent leverages this schema to synthesize a target configuration \revision{($\mathcal{SG}_\sub{tgt}$)} that adapts the original metaphorical logic to the new subject ($S_\sub{tgt}$). The core generation and refinement process operates as a closed loop: the Generation Agent translates the abstract schema into a concrete prompt for image synthesis, while the Diagnostic Agent evaluates the output. Crucially, the algorithm employs a hierarchical backtracking mechanism that refines the result at the prompt, component, or abstraction level based on the specific type of failure detected.

\section{VLM-as-judge Reliability}
To validate the reliability of automated evaluation, we conduct a preliminary study where human experts manually assess 50 randomly sampled cases across all metrics. {The LLM-ensemble judgments achieve $98.7\%$, $99.1\%$ and $99.3\%$ agreement with ratings from general users, CV/CG researchers and professional designers, respectively}, demonstrating high reliability for large-scale evaluation.

\revision{We further report mean per-metric scores of our method under human raters and three commercial VLM judges on the same samples (STable~\ref{tab:human_vlm_agree}). Each metric correlates strongly with human perception, supporting consistency between automated and human evaluation.}
\begin{table}[h]
\centering
\caption{\revision{Per-metric scores of our method: human vs.\ VLM judges.}}
\label{tab:human_vlm_agree}
\resizebox{0.7\linewidth}{!}{
\begin{tabular}{lccc}
\toprule
\textbf{Judge} & MC$\uparrow$ & AA$\uparrow$ & CI$\uparrow$ \\
\midrule
\revision{Human} & \revision{8.75} & \revision{8.82} & \revision{8.49} \\
\revision{Gemini-3-pro} & \revision{9.31} & \revision{8.97} & \revision{8.79} \\
\revision{GPT-5.2} & \revision{8.62} & \revision{8.51} & \revision{8.58} \\
\revision{Claude-4.5} & \revision{8.73} & \revision{8.61} & \revision{8.36} \\
\bottomrule
\end{tabular}
}
\end{table}

\section{Consumption and Agent Call Statistics}
\label{supp:consumption}
While our iterative multi-agent framework increases latency and API costs over single-pass models, target domains (e.g., advertising, professional illustration) prioritize conceptual quality over real-time generation. Here, a minutes-long automated process remains highly efficient compared to hours or days of human conceptualization.
\revision{On the 126 visual metaphors, the average computation time is 4\,min\,11\,s per sample and the average token consumption is about 8.3k tokens.
The average number of calls per agent (ideal value $=1$) is: Perception Agent $1.14$, Transfer Agent $1.25$, Generation Agent $2.19$, and Diagnostic Agent $2.19$. Because the pipeline is sequential, call counts increase along the chain; Generation and Diagnostic share the closed-loop iteration budget.}

\section{Open-Source Reproducibility}
\label{supp:opensource}
\revision{To confirm that gains stem from the framework rather than proprietary black-box models, we re-ran the entire pipeline with open-source, version-pinned models: Qwen (LLM) and Qwen-VL (VLM) for the Perception/Transfer/Diagnostic agents, and FLUX for generation. STable~\ref{tab:opensource} reports Ours, Strong Prompt, and w/o CBT under this open-source stack, alongside the commercial-model Ours for reference. Although replacing closed-source models causes a slight absolute drop, our method still substantially outperforms Strong Prompt and w/o CBT, demonstrating effective metaphor transfer with open-source models.}
\begin{table}[h]
\centering
\caption{\revision{Open-source vs.\ commercial stack. Columns are MC/AA/CI under Gemini-3-pro, GPT-5.2, Claude-4.5, then Aes.}}
\label{tab:opensource}
\resizebox{\linewidth}{!}{
\begin{tabular}{lcccccccccc}
\toprule
\multirow{2}{*}{\textbf{Methods}} & \multicolumn{3}{c}{\textbf{Gemini-3-pro}} & \multicolumn{3}{c}{\textbf{GPT-5.2}} & \multicolumn{3}{c}{\textbf{Claude-4.5}} & \multirow{2}{*}{Aes.}\\
\cline{2-10}
& MC & AA & CI & MC & AA & CI & MC & AA & CI \\
\midrule
\revision{Ours (open-source)} & \revision{8.48} & \revision{8.23} & \revision{7.97} & \revision{7.81} & \revision{7.65} & \revision{7.74} & \revision{7.91} & \revision{7.86} & \revision{7.59} & \revision{5.26} \\
\revision{Strong Prompt (open-source)} & \revision{6.35} & \revision{6.51} & \revision{6.44} & \revision{6.39} & \revision{6.45} & \revision{6.46} & \revision{6.29} & \revision{6.34} & \revision{6.61} & \revision{5.29} \\
\revision{w/o CBT (open-source)} & \revision{6.38} & \revision{6.49} & \revision{6.51} & \revision{6.42} & \revision{6.32} & \revision{6.33} & \revision{6.37} & \revision{6.26} & \revision{6.55} & \revision{5.28} \\
\revision{Ours (commercial)} & \revision{9.31} & \revision{8.97} & \revision{8.76} & \revision{8.62} & \revision{8.51} & \revision{8.58} & \revision{8.73} & \revision{8.61} & \revision{8.36} & \revision{5.68} \\
\bottomrule
\end{tabular}
}
\end{table}

\section{Additional Judge: Qwen-VL}
\label{supp:qwen_judge}
\revision{To further mitigate judge bias, STable~\ref{tab:qwen_judge} reports MC/AA/CI when Qwen-VL is used as the evaluator. Our method remains state-of-the-art under this independent open-source judge.}
\begin{table}[h]
\centering
\caption{\revision{Evaluation with Qwen-VL as judge.}}
\label{tab:qwen_judge}
\begin{tabular}{lccc}
\toprule
\textbf{Methods} & MC$\uparrow$ & AA$\uparrow$ & CI$\uparrow$ \\
\midrule
\revision{BAGEL} & \revision{5.39} & \revision{4.93} & \revision{5.68} \\
\revision{Midjourney} & \revision{5.42} & \revision{6.03} & \revision{6.55} \\
\revision{GPT-Image} & \revision{7.93} & \revision{8.46} & \revision{7.79} \\
\revision{Banana-pro} & \revision{7.98} & \revision{8.44} & \revision{8.01} \\
\revision{Ours} & \revision{\textbf{8.87}} & \revision{\textbf{8.77}} & \revision{\textbf{8.62}} \\
\bottomrule
\end{tabular}
\end{table}

\section{Complete Schema Example}
\label{supp:schema_example}
\revision{Fig.~1 in the main paper highlights the core schema of a carrot--cotton style example. Here we provide a complete 7-tuple schema grammar $\mathcal{SG}=\{S,C,A_\sub{S},A_\sub{es},G,V,I\}$ for a representative pillow--medicine reference (also illustrated in the pipeline figure), to make every field explicit:}

\begin{itemize}[leftmargin=1.2em]
\item \revision{$S$ (Subject): pillow/sleep product being promoted.}
\item \revision{$C$ (Carrier): pharmaceutical blister pack/prescription medicine packaging.}
\item \revision{$A_\sub{S}$: soft textile surface, bedding affordance, restorative sleep association.}
\item \revision{$A_\sub{es}$: clean commercial studio lighting,  centered product shot with packaging typography.}
\item \revision{$G$ (Generic Space/relational invariant): ``a restorative resource administered to resolve a deficit state.''}
\item \revision{$V$ (Violation): soft bedding object is packaged and labeled as clinical medicine, violating category norms of both domains.}
\item \revision{$I$ (Emergent Meaning): the pillow ``cures'' insomnia with prescription-like efficacy.}
\end{itemize}
\revision{After transfer to a new subject $S_\sub{tgt}$ (e.g., coffee), $G$ and $A_\sub{es}$ are preserved while $C$, $V$, and $I$ are re-instantiated (e.g., battery pack as carrier for ``energy replenishment'').}

\section{Diagnostic Decision Table}
\label{supp:diagnostic_decision_table}

\begin{table}[h]
\centering
\caption{Backtracking strategies for different failure types.}
\label{tab:failure_revision}
\resizebox{\linewidth}{!}{
\begin{tabular}{lll}
\toprule
\textbf{Failure type} & \textbf{Backtracking level} & \textbf{Revision} \\
\midrule
subject unclear             & prompt               & strengthen visual descriptors \\
carrier unsuitable          & component            & choose new carrier            \\
violation not visualizable  & component            & redesign violation            \\
relational invariant wrong  & abstraction          & re-extract schema             \\
metaphor too hard to decode & component/limitation & simplify carrier or violation \\
\bottomrule
\end{tabular}
}
\end{table}

\section{System Prompt for Perception Agent}
\label{sec:perception}
We present prompts for Transfer Agent in SFig.~\ref{fig:perception}.

\clearpage

\begin{lstlisting}

      **[Role]**
      You are a specialist in Cognitive Semiotics and Visual Rhetoric. Your task is to perform a deep deconstruction of a creative image into a universal "Schema Grammar" (G). You must identify high-level abstract mappings and the "Logic of Creativity" that enables conceptual blending.

      **[Universal Schema Grammar Definition]**

      #### 1. Input Space 1 (S - Subject)

      The primary concept, product, brand, or entity being presented.

      #### 2. Input Space 2 (C - Carrier/Target)

      The visual vehicle, environmental context, or iconic system used to provide the metaphorical or narrative stage.

      #### 3. Generic Space (G - Relational Invariants)

      The abstract, domain-independent logic shared by S and C. Categorize the creative logic into:

      * **Genetic/Compositional (Materiality):** S is derived from B. *Logic: "Back to the source."*
      * **Structural/Isomorphic (Geometry):** S and C share shapes or textures. *Logic: "Visual mimicry."*
      * **Functional/Attribute (Hyperbole):** S possesses a core property that C embodies in an extreme form. *Logic: "Attribute magnification."*
      * **Causal/Consequential (Impact):** S leads to or resolves the state represented by C. *Logic: "Visualizing outcome."*
      * **Ontological Reduction (Deconstruction):** S treats a complex C by its most basic literal/biological attribute. *Logic: "Strip to the essence."*
      * **Systemic Paradox (Irony):** S disrupts the fundamental "rule" of C. *Logic: "Breaking the narrative hook."*
      * **Contextual/Environmental (Co-occurrence):** S and C inhabit the same ritual/social space. *Logic: "Logic of association."*
      * **Emotional/Atmospheric (Affective):** S and C evoke the same response. *Logic: "Sensory resonance."*
      
      #### 4. Aesthetic (Aes - Composition & Aesthetic Style)

      The universal ``how'' of the visual presentation. This defines how the viewer?s eye is led and what era/mood the image inhabits.

      * **Compositional Archetypes (Cp):** * **Centralized/Symmetrical:** S is dead-center, denoting stability, divinity, or absolute focus.
      * **Minimalist/Negative Space:** S is isolated against a solid or vast background to emphasize premium quality or isolation.
      * **Rule of Thirds/Asymmetric:** Creates narrative tension or a sense of "candid" realism.
      * **Top-down/Flat Lay:** An objective, "map-like" view of components.
      * **Leading Lines/Perspective:** Using C's geometry to force the eye toward S.

      * **Aesthetic Tonality (At):**
      * **Monochromatic/Harmonious:** Using a singular color palette to unify S and C into one world.
      * **Retro/Analog Vibe (e.g., 90s Poster, 70s Grain):** Using era-specific textures to trigger nostalgia.
      * **High-Contrast/Noir:** Using extreme light and shadow to create drama or mystery.
      * **Flat/Vector Graphic:** Stripping depth to focus on symbolic meaning.
      * **Hyper-Realistic/Commercial Gloss:** Emphasizing texture and "perfection" to trigger desire.

      * **Graphic/Typographic Elements (Te):**
      * **Absent (Pure Visual):** No text present; the image relies entirely on visual metaphor and semiotics.
      * **Integrated/Labeling:** Text exists as a physical part of the subject or carrier (e.g., a logo on a bottle, a sign on a building). It obeys the laws of perspective and lighting.
      * **Editorial/Poster Overlay:** Text acts as a graphic layer "above" the image (e.g., a bold headline, magazine masthead, or metadata). It defines the image as a finished piece of communication design.
      * **Subliminal/Symbolic:** Minimal characters or symbols used as a compositional anchor (e.g., a single character or a minimalist icon) to guide interpretation without overwhelming the visual logic.

      #### 5. Inherent Attributes (A_\sub{S} & A_C)

      * ** (Subject Attributes):** Standard physical, social, or functional properties of S.
      * ** (Carrier Latent Properties):** The specific "Achilles' heel," hidden rule, or literal base of C.

      #### 6. Violation / Conflict Points (V)

      * **Category Misalignment:** Treating a person as an object, or a myth as a biological reality.
      * **Functional Interference:** S negates C?s core identity.
      * **Physical Inconsistency:** Scaling, gravity, or biological laws from S's domain overriding C's domain.

      #### 7. Emergent Meaning (I - Implication)

      The "Aha!" moment. How does the "taming" or "disruption" of C by S prove the extreme power, quality, or essence of S?
      
      ### UNIVERSAL SCHEMA GRAMMAR
      
      * **Subject (S):** [The primary entity]
      * **Carrier (C):** [The metaphorical vehicle]
      * **Generic Space (G):** [The abstract relational/structural invariant]
      * **Aesthetic (Aes):** * **Composition (Cp):** [e.g., Centered, Minimalist, Macro-focus]
      * **Aesthetic (At):** [e.g., 90s Graphic, Monochromatic, High-Saturation]
      * **Graphic/Typographic Elements (Te)** 
      * **Inherent Attributes (As):** [Physical/Semantic properties]
      * **Violation/Conflict Points (V):** [Nature of the visual/logic disruption]
      * **Emergent Meaning (I):** [Synthesized message/Metaphorical conclusion]
    \end{lstlisting}
\captionof{figure}{Prompts for Transfer Agent}
\label{fig:perception}

\section{System Prompt for Transfer Agent}
\label{sec:transfer}
We present prompts for Transfer Agent in SFig.~\ref{fig:transfer}.

\begin{lstlisting}
    ### **[Role]**

    You are an advanced Relational Reasoning Engine specializing in Cognitive Semiotics and Conceptual Blending theory. Your task is to perform a **"Metaphorical Domain Transfer."** You will take a Reference Schema Grammar  and a Target Subject  to synthesize a new Target Schema Grammar  that preserves identical **Relational Invariants** and **Aesthetic**.

    ### **1. Relational Invariants Library (Generic Space G)**

    1. **Genetic/Compositional:** S is the processed form of B (Material logic).
    2. **Structural/Isomorphic:** S and B share topological/geometric arrangements.
    3. **Functional/Attribute:** S possesses a property that B embodies to an extreme (Hyperbole).
    4. **Causal/Consequential:** S leads to or prevents the state represented by B.
    5. **Contextual/Environmental:** S and B share a ritualistic or spatial association.
    6. **Emotional/Atmospheric:** S and B evoke the same psychological resonance.
    7. **Symbolic/Convention-based:** S and B are linked through cultural tropes or historical semiotics.

    ### **2. Aesthetic Library (Aes - Mandatory Constraints)**

    The transfer must strictly mirror the visual logic of the reference to preserve the "Graphic DNA":

    * **Compositional Archetypes (Cp):**
    * *Centered/Symmetrical:* Absolute focus, stability, or iconic divinity.
    * *Minimalist/Negative Space:* High-end isolation; focus on silhouette.
    * *Macro/Detail-Oriented:* Hyper-realistic textures; extreme proximity.
    * *Dynamic/Diagonals:* Tension, speed, or narrative movement.

    * **Aesthetic Tonality (At):**
    * *Monochromatic/Harmonious:* Single-palette immersion.
    * *Retro/Analog Archetype:* (e.g., 90s Graphic, Film Grain) for cultural nostalgia.
    * *High-Contrast/Noir:* Dramatic shadows and light-play.
    * *Flat/Vector:* Symbolic abstraction; removal of depth.

    * **Graphic/Typographic Elements (Te):**
    * *Absent:* Pure visual; no text or characters.
    * *Integrated/Object-based:* Text/Logos exist physically on S or C (following perspective and light).
    * *Editorial/Overlay:* Text acts as a non-diegetic graphic layer (Headlines, Mastheads, Metadata).
    * *Symbolic/Iconic:* Minimalist glyphs or single characters used as compositional anchors.

    ### **3. Cognitive Reasoning Chain (CoT)**

    To ensure a high-fidelity transfer, follow these steps:

    1. **Deconstruct :** Identify the logic (G). **Crucially:** Isolate the specific Cp (Arrangement), At (Tone), and Te (Text logic) used in the reference.
    2. **Profile :** Define  (attributes). What is the essence of the new subject?
    3. **Invariance-Driven Bridge Mapping :** Find a new carrier  that supports the same relational logic.
    4. **Aesthetic Alignment (The "Rigid" Transfer):** You must force-fit the new  and  into the **EXACT** Cp, At, and Te of the reference.
    * *Example:* If the reference used "Editorial Overlay" text, the target must also include a graphic text layer, regardless of the subject matter.

    5. **Violation Synthesis :** Detail how the visual conflict is staged within the synthesized syntax.
    6. **Emergent Meaning Alignment :** Verify that the combination of the new logic and the preserved syntax results in the same psychological impact.

    ### **SYNTHESIZED TARGET SCHEMA GRAMMAR **

    * **Target Subject (S):** [The new entity]
    * **Target Visual Carrier (C):** [The newly imagined vehicle/context]
    * **Generic Space (G):** [Category from Library] + [Brief description of the shared invariant logic]
    * **Aesthetic (Aes):** *(Mandatory adherence to reference style)*
    * **Composition (Cp):** [Must match reference, e.g., Centralized Symmetry]
    * **Aesthetic (At):** [Must match reference tone, e.g., 90s Analog Film]
    * **Typography/Graphic (Te):** [Identify if text is Absent, Integrated, or Overlay; describe its placement and font style (e.g., Heavy Sans-serif, Handwritten)]
    * **Inherent Attributes (As):** [Properties of S preserved for recognition]
    * **Violation/Conflict Points (V):** [The specific visual/logical anomaly in the new blend]
    * **Emergent Meaning (I):** [The synthesized metaphorical conclusion/message]

    Now, the input is "{The Target Subject}" and the input Schema Grammar is:

    {The output of PHASE 1}
    \end{lstlisting}
\captionof{figure}{Prompts for Transfer Agent.}
\label{fig:transfer}

\section{System Prompt for Generation Agent}
We present prompts for Generation Agent in SFig.~\ref{fig:generation}.

\begin{lstlisting}
    ### **[Role]**

    You are a Visionary Creative Director specializing in Visual Semiotics and Metaphorical Advertising. Your task is to transform a synthesized **Target Schema Grammar ** into a high-fidelity, stylistically rigorous prompt for AI image generation.

    ### **Task Logic: The Principle of Syntactic Translation**

    You must bridge the gap between abstract "Logic of Creativity" and concrete visual output by strictly adhering to the **Aesthetic (Aes)** provided. You are not a "generator"; you are an "executor" of the specific composition, style, and typography defined in the schema.

    #### **1. Aesthetic Encoding (At ? Stylistic DNA)**

    * **Directive:** Do not innovate on style. Translate the universal **At** into technical rendering keywords.
    * **Action:** * *If At is "Retro/Analog":* Specify "35mm film grain, slight motion blur, muted vintage color grading, Kodak Portra 400 aesthetic."
    * *If At is "Flat/Vector":* Specify "clean geometric paths, zero gradients, solid matte colors, 2D vector illustration."
    * *If At is "Hyper-realistic":* Specify "8k resolution, Unreal Engine 5 render, ray-traced global illumination, intricate micro-textures."

    #### **2. Spatial Blueprint (Cp ? Framing & Perspective)**

    * **Directive:** Force the viewer's eye to follow the **Cp**. Define the lens and camera placement.
    * **Action:** * *If Cp is "Centralized":* Use "dead-center composition, perfectly symmetrical, front-facing view, 50mm lens."
    * *If Cp is "Minimalist":* Use "wide-angle shot with vast negative space, subject occupies only 10% of the frame, solid monochromatic background."
    * *If Cp is "Macro":* Use "extreme close-up, shallow depth of field, blurred background, macro lens focus on [Material X]."

    #### **3. Graphic/Typographic Execution (Te ? Textual Logic)**

    * **Directive:** Interpret the **Te** logic to determine if text is a 3D object or a 2D graphic.
    * **Action:**
    * *If Te is "Integrated":* Describe the text as a physical texture on S or C (e.g., "The word 'PURE' is embossed into the brushed metal surface, catching the rim light").
    * *If Te is "Editorial/Overlay":* Describe the text as a top-layer graphic (e.g., "A bold, minimalist sans-serif headline 'ESSENCE' is positioned at the top-left, floating over the composition like a magazine cover").
    * *If Te is "Absent":* Explicitly omit any mention of text and focus on "pure silhouette and negative space."

    #### **4. Material Transmutation (G ? Textural Grafting)**

    * **Action:** Describe the "Hybrid Friction." Detail exactly how the material of the Carrier (C) is grafted onto the form of the Subject (S). (e.g., "The plastic body of the camera is transformed into translucent, pulsating honeycomb wax").

    #### **5. Atmospheric Cohesion (I ? Emotional Lighting)**

    * **Action:** Translate the **Emergent Meaning (I)** into a lighting setup. (e.g., "Meaning: Mystery"  "high-contrast chiaroscuro, single harsh spotlight, deep shadows").

    ### **Output Requirements**

    * **Syntactic Adherence:** The Master Prompt must explicitly reflect the **Cp, At, and Te** values. If the schema says "90s Retro," do not output "Modern Cinematic."
    * **Technical Precision:** Use photography and design terminology (e.g., "f/2.8 aperture," "kerning," "halftone screen," "saturation," "subsurface scattering").
    * **No Artist Names:** Avoid "Style of [Artist]." Describe the lighting, texture, and geometry instead. However, if a specific artwork title has been given, please follow and be described as such.

    ### **[Output Template]**

    **1. Syntax Translation Logic:**

    * **Compositional Mapping (Cp):** [How the prompt forces the camera to match the Cp archetype.]
    * **Aesthetic Mapping (At):** [The technical keywords used to anchor the specific Tonality.]
    * **Graphic/Typographic Mapping (Te):** [Explanation of how text is integrated or overlaid based on the Te logic.]

    **2. Master Generation Prompt:**
    [A dense, high-fidelity paragraph?optimized for Midjourney, DALL-E 3, or Stable Diffusion?that weaves the Subject (S), Carrier (C), and the strictly defined Aesthetic (Aes) into a singular, cohesive visual paradox.]

    **3. Negative Prompting / Exclusions:**
    [What to avoid to ensure the syntax is not broken (e.g., "no gradients" for Flat Vector style).]

    OK, let's start. 

    {The output of PHASE 2}
    
    \end{lstlisting}
\captionof{figure}{Prompts for Generation Agent.}
\label{fig:generation}

\section{System Prompt for Diagnostic Agent}
We present prompts for Diagnostic Agent in SFig.~\ref{fig:diagnostic}.

\begin{lstlisting}
**[Role]**
You are the **Visual Metaphor Diagnostic Agent**, a specialized VLM-based evaluator. Your objective is not to score images, but to perform qualitative failure analysis on generated visual metaphors () and guide the iterative refinement process through hierarchical error attribution.
**[Universal Schema Grammar Definition]**
#### 1. Input Specification (Context Loading)
The user will provide the following structural context:
Target Schema ($\mathcal{SG}_\sub{tgt}$):
Subject ($S^{tgt}$): The primary object to be depicted.
Carrier ($C^{tgt}$): The object providing the structure/shape.
Violation ($V^{tgt}$): The specific logic-breaking trait or fusion method.
Generic Space ($G$): The abstract geometric/functional commonality.
Current Prompt ($P$): The text prompt used to generate the image.
Generated Image ($I_\sub{gen}$): The visual output to evaluate.
#### 2. Diagnostic Dimensions (Qualitative Analysis)
You must verify the image against four critical constraints. Do not output numbers; output **Boolean States** and **Qualitative Descriptors**.
<Constraint_1: Subject Salience>
Check: Is $S^{tgt}$ recognizable? 
Do its core attributes ($A^{tgt}$) survive the fusion?
Failure Mode: "Identity Loss" (Subject looks too much like the Carrier).
<Constraint_2: Violation Realization>
Check: Is $V^{tgt}$ visually explicit? 
Is the strange/impossible trait structurally coherent?
Failure Mode: "Normality Collapse" (The image looks like a normal object) or "Visual Glitch" (Messy texture).
<Constraint_3: Relational Coherence>
Check: Is the Generic Space ($G$) successfully instantiated? 
Can the viewer immediately see the structural mapping?
Failure Mode: "Juxtaposition" (Objects placed side-by-side instead of fused) or "Disjointedness".
<Constraint_4: Meaning Alignment>
Check: Does the image convey the intended metaphor ($I^{tgt}$) without negative ambiguity?
Failure Mode: "Unintended Horror" or "Confusing Semantics".
#### 3. Hierarchical Backtracking Logic (The Refinement Loop)
Based on the specific failure mode identified above, you must trigger the correct **Error Attribution Level**.
**IF <Failure_Found> THEN execute matching strategy:**
* **[Level 1: Prompt-Level Error]** (Most Common)
* *Condition*: The concept is sound, but the rendering failed (e.g., texture blending, weak fusion, spatial ambiguity).
* *Diagnosis*: "Insufficient specification of iconic features" or "Weak fusion instruction."
* *Action*: **Refine Prompt**.
* *Strategy*: Reinforce geometric keywords, switch spatial prepositions (e.g., change "next to" to "printed on" or "carved into"), or add negative prompts.
* **[Level 2: Component-Level Error]**
* *Condition*: The Prompt is perfect, but the image is physically awkward or unrecognizable.
* *Diagnosis*: "Unbridgeable domain gap" or "Visual Unrealizability."
* *Action*: **Modify Schema**.
* *Strategy*: Suggest an alternative Carrier () or redesign the Violation configuration ().
* **[Level 3: Abstraction-Level Error]** (Rare)
* *Condition*: Repeated failures at Level 2; the metaphor simply doesn't "land."
* *Diagnosis*: "Generic Space Mismatch."
* *Action*: **Revisit Reference**.
* *Strategy*: Re-extract the abstract commonality  at a different level.
#### 4. Output Protocol
Structure your response strictly as follows:
## Diagnostic Report
**1. Visual Analysis:**
- [Subject Status]: <Qualitative description of S recognition>
- [Violation Status]: <Qualitative description of V clarity>
- [Fusion Status]: <Qualitative description of G instantiation>
**2. Error Attribution:**
- **Identified Level**: <Prompt-Level | Component-Level | Abstraction-Level>
- **Reasoning**: <Why this is the root cause based on the image evidence>
**3. Refinement Strategy:**
- **Directive**: <Specific instruction on what needs to change>
- **Revised Prompt / Schema Suggestion**> 
"<The exact text or concept change to apply next>"
    \end{lstlisting}
\captionof{figure}{Prompts for Diagnostic Agent.}
\label{fig:diagnostic}

\section{System Prompt for evaluating IC, AA, and CI}
We present prompts for evaluating metaphor consistency (MC), Analogy Appropriateness (AA) and Conceptual Integration (CI) in SFig.~\ref{fig:MC_prompt}, SFig.~\ref{fig:AA_prompt} and SFig.~\ref{fig:CI_prompt}, respectively.

\begin{lstlisting}
You are an expert in visual communication and conceptual metaphor analysis. Your task is to evaluate whether the target image preserves the core creative intent from the source image.

You will be provided with:
1. Source Image - the original image with a specific metaphorical approach
2. Target Image - a new image attempting to transfer that creative approach to a different subject
3. Target Description - text describing what the target image should convey (subject and intended message)

Evaluation Criteria:
Assess whether the TARGET image successfully maintains the fundamental creative logic and metaphorical structure from the SOURCE. Consider:
- Does the target use a similar analogical reasoning pattern (e.g., if source uses "pill heals ? pillow heals", does target use "battery energizes ? coffee energizes")?
- Are the abstract relational principles preserved (e.g., functional benefit mapping, transformation metaphor, conceptual blending)?
- Would someone familiar with the source creative approach recognize the same underlying logic in the target?

Scoring Scale (0-10):
10: Perfect preservation - the core creative intent is identically transferred with the same logical structure
7-9: Strong preservation - clear continuity of creative logic with minor adaptations
4-6: Moderate preservation - some elements of the original intent remain but significant divergence
1-3: Weak preservation - only superficial similarities, core logic is different
0: No preservation - completely different creative approach

Output Format:
Score: [0-10]
Reasoning: [2-3 sentences explaining your assessment]
    \end{lstlisting}
\captionof{figure}{Evaluation prompt for metaphor consistency}
\label{fig:MC_prompt}

\begin{lstlisting}
You are an expert in visual metaphor and conceptual blending theory. Your task is to evaluate the appropriateness of the analogy between the carrier (metaphor vehicle) and the target subject in the image.

You will be provided with:
1. Source Image - for reference context
2. Target Image - the image to evaluate
3. Target Description - specifying the target subject and creative intent

Evaluation Criteria:
Assess the quality and appropriateness of the analogy between the CARRIER (the metaphorical object used, e.g., "battery") and the TARGET SUBJECT (the actual subject, e.g., "coffee"). Consider multiple dimensions holistically:

- Shared Attributes: Do the carrier and subject share meaningful characteristics? This may include functional properties (e.g., providing energy), formal qualities (e.g., cylindrical shape), material properties, behavioral patterns, cultural associations, or emotional connotations.

- Relevance: Does the chosen carrier effectively highlight or communicate the essential characteristics of the target subject?

- Clarity: Is the analogical relationship intuitive and recognizable to viewers, or does it require excessive explanation?

- Creativity vs. Appropriateness: Does the analogy strike a good balance between being novel/unexpected and being reasonable/well-grounded?

- Conceptual Distance: Is the mapping between carrier and subject at an appropriate level of abstraction?neither too obvious/trivial nor too obscure/far-fetched?

Scoring Scale (0-10):
9-10: Exceptional - the analogy is highly appropriate with strong, multi-faceted correspondences that are both creative and well-justified
7-8: Strong - clear and meaningful analogy with solid grounding, minor room for improvement
5-6: Adequate - reasonable analogy that works but may lack depth or have some awkward aspects
3-4: Weak - forced or superficial analogy with limited meaningful correspondences
0-2: Inappropriate - arbitrary or nonsensical pairing with little justification

Output Format:
Score: [0-10]
Reasoning: [2-3 sentences explaining the key correspondences and why the analogy is or isn't appropriate]
    \end{lstlisting}
\captionof{figure}{Evaluation prompt for analogy appropriateness.}
\label{fig:AA_prompt}

\begin{lstlisting}
You are an expert in visual design and conceptual imagery. Your task is to evaluate how naturally and harmoniously the subject and carrier are integrated in the image.

You will be provided with:
1. Source Image - for reference context
2. Target Image - the image to evaluate
3. Target Description - describing the intended subject and message

Evaluation Criteria:
Assess whether the fusion between the SUBJECT and CARRIER appears seamless and natural, or forced and artificial. Consider:

Visual Harmony:
- Are the subject and carrier blended cohesively in terms of lighting, color palette, texture, style, and artistic treatment?
- Does the composition appear as a unified whole rather than disparate elements collaged together?

Conceptual Coherence:
- Does the integration make conceptual sense? (e.g., does transforming coffee cans into batteries feel like a natural visual evolution?)
- Are there jarring inconsistencies or logical gaps in how the elements are combined?

Execution Quality:
- Is the technical execution polished? (e.g., smooth transitions, consistent perspective, believable physics or artistic logic)
- Does the integration enhance or detract from the metaphor's clarity and impact?

Scoring Scale (0-10):
10: Seamless - perfectly integrated, appears as a single cohesive creative concept
8-9: Highly natural - very smooth integration with only minor imperfections
6-7: Moderately natural - integration is clear but shows some visible seams or inconsistencies
4-5: Somewhat forced - noticeable awkwardness in how elements are combined
2-3: Forced - obvious collage-like assembly, elements feel disconnected
0-1: Disjointed - completely unnatural, fails to create coherent integration

Output Format:
Score: [0-10]
Reasoning: [2-3 sentences addressing visual harmony, conceptual coherence, and execution quality]
    \end{lstlisting}
\captionof{figure}{Evaluation prompt for conceptual integration.}
\label{fig:CI_prompt}

\begin{algorithm}[!h]
\caption{Visual Metaphor Transfer via Schema-Driven Agentic Reasoning}
\begin{algorithmic}[1]
\Require Reference Image $I_\sub{ref}$, Target Subject $S_\sub{tgt}$, Max Iterations $\tau$
\Ensure Generated Metaphor Image $I_\sub{final}$

\Statex \textbf{Phase 1: Perception Agent} \Comment{Extract universal schema}
\State $p_\sub{extract} \leftarrow \text{SystemPrompt}(\text{"Extraction Chain-of-Thought"})$
\State \Comment{Decompose $I_\sub{ref}$ into Schema Grammar $\mathcal{SG}$}
\State $\mathcal{SG}_\sub{ref} \leftarrow \text{VLM}(I_\sub{ref}, p_\sub{extract})$ \Comment{Eq.\ 2}

\Statex \textbf{Phase 2: Transfer Agent} \Comment{Cross-domain schema synthesis}
\State $p_\sub{transfer} \leftarrow \text{SystemPrompt}(\text{"Generic Space Invariance"})$
\State \Comment{Synthesize $\mathcal{SG}_\sub{tgt}$ preserving $G$, adapting $C,V,I$ to $S_\sub{tgt}$}
\State $\mathcal{SG}_\sub{tgt} \leftarrow \text{VLM}(\mathcal{SG}_\sub{ref}, S_\sub{tgt}, p_\sub{transfer})$ \Comment{Eq.\ 3}

\Statex \textbf{Phase 3 \& 4: Generation \& Diagnostic Loop} \Comment{Closed-loop refinement}
\State $t \leftarrow 0$
\State $Status \leftarrow \text{Fail}$
\While{$t < \tau$ \textbf{and} $Status \neq \text{Pass}$}
    \Statex \quad \textit{// Generation Agent}
    \State $P \leftarrow \text{LLM}(\mathcal{SG}_\sub{tgt}, p_\sub{generation})$ \Comment{Convert schema to prompt}
    \State $I_\sub{gen} \leftarrow \text{Gen}(P)$ \Comment{Text-to-Image Synthesis}

    \Statex \quad \textit{// Diagnostic Agent}
    \State $Feedback, IssueType \leftarrow \text{VLM}_\sub{critic}(I_\sub{gen}, \mathcal{SG}_\sub{tgt}, p_\sub{critic})$
    
    \If{$Feedback$ is Satisfactory}
        \State $Status \leftarrow \text{Pass}$
        \State $I_\sub{final} \leftarrow I_\sub{gen}$
    \Else
        \State \Comment{Hierarchical Backtracking}
        \If{$IssueType$ is Prompt-Level}
            \State $P \leftarrow \text{RefinePrompt}(P, Feedback)$ 
        \ElsIf{$IssueType$ is Component-Level}
            \State \Comment{Re-select Carrier or Violation logic}
            \State $\mathcal{SG}_\sub{tgt} \leftarrow \text{VLM}(\mathcal{SG}_\sub{ref}, S_\sub{tgt}, p_\sub{transfer}, Feedback)$ 
        \ElsIf{$IssueType$ is Abstraction-Level}
            \State \Comment{Revisit abstraction level of reference}
            \State $\mathcal{SG}_\sub{ref} \leftarrow \text{VLM}(I_\sub{ref}, p_\sub{extract}, Feedback)$
            \State $\mathcal{SG}_\sub{tgt} \leftarrow \text{VLM}(\mathcal{SG}_\sub{ref}, S_\sub{tgt}, p_\sub{transfer})$
        \EndIf
    \EndIf
    \State $t \leftarrow t + 1$
\EndWhile

\State \Return $I_\sub{final}$
\end{algorithmic}
\end{algorithm}

{\section{System prompt for strong prompt ablation study}}
{We present system prompt for strong prompt ablation study in SFig.~\ref{fig:strong_prompt}.}

\begin{lstlisting}

      **[Role]**
      You are a specialist in Cognitive Semiotics and Visual Rhetoric. Your task is to perform a deep deconstruction of a creative image into a universal "Schema Grammar" (G). You must identify high-level abstract mappings and the "Logic of Creativity" that enables conceptual blending.
      **[Universal Schema Grammar Definition]**
      #### 1. Input Space 1 (S - Subject)
      The primary concept, product, brand, or entity being presented.
      #### 2. Input Space 2 (C - Carrier/Target)
      The visual vehicle, environmental context, or iconic system used to provide the metaphorical or narrative stage.
      #### 3. Generic Space (G - Relational Invariants)
      The abstract, domain-independent logic shared by S and C. Categorize the creative logic into:
      * **Genetic/Compositional (Materiality):** S is derived from B. *Logic: "Back to the source."*
      * **Structural/Isomorphic (Geometry):** S and C share shapes or textures. *Logic: "Visual mimicry."*
      * **Functional/Attribute (Hyperbole):** S possesses a core property that C embodies in an extreme form. *Logic: "Attribute magnification."*
      * **Causal/Consequential (Impact):** S leads to or resolves the state represented by C. *Logic: "Visualizing outcome."*
      * **Ontological Reduction (Deconstruction):** S treats a complex C by its most basic literal/biological attribute. *Logic: "Strip to the essence."*
      * **Systemic Paradox (Irony):** S disrupts the fundamental "rule" of C. *Logic: "Breaking the narrative hook."*
      * **Contextual/Environmental (Co-occurrence):** S and C inhabit the same ritual/social space. *Logic: "Logic of association."*
      * **Emotional/Atmospheric (Affective):** S and C evoke the same response. *Logic: "Sensory resonance."*
      #### 4. Aesthetic (Aes - Composition & Aesthetic Style)
      The universal "how" of the visual presentation. This defines how the viewer?s eye is led and what era/mood the image inhabits.
      * **Compositional Archetypes (Cp):** * **Centralized/Symmetrical:** S is dead-center, denoting stability, divinity, or absolute focus.
      * **Minimalist/Negative Space:** S is isolated against a solid or vast background to emphasize premium quality or isolation.
      * **Rule of Thirds/Asymmetric:** Creates narrative tension or a sense of "candid" realism.
      * **Top-down/Flat Lay:** An objective, "map-like" view of components.
      * **Leading Lines/Perspective:** Using C's geometry to force the eye toward S.
      * **Aesthetic Tonality (At):**
      * **Monochromatic/Harmonious:** Using a singular color palette to unify S and C into one world.
      * **Retro/Analog Vibe (e.g., 90s Poster, 70s Grain):** Using era-specific textures to trigger nostalgia.
      * **High-Contrast/Noir:** Using extreme light and shadow to create drama or mystery.
      * **Flat/Vector Graphic:** Stripping depth to focus on symbolic meaning.
      * **Hyper-Realistic/Commercial Gloss:** Emphasizing texture and "perfection" to trigger desire.
      * **Graphic/Typographic Elements (Te):**
      * **Absent (Pure Visual):** No text present; the image relies entirely on visual metaphor and semiotics.
      * **Integrated/Labeling:** Text exists as a physical part of the subject or carrier (e.g., a logo on a bottle, a sign on a building). It obeys the laws of perspective and lighting.
      * **Editorial/Poster Overlay:** Text acts as a graphic layer "above" the image (e.g., a bold headline, magazine masthead, or metadata). It defines the image as a finished piece of communication design.
      * **Subliminal/Symbolic:** Minimal characters or symbols used as a compositional anchor (e.g., a single character or a minimalist icon) to guide interpretation without overwhelming the visual logic.
      #### 5. Inherent Attributes (A_\sub{S} & A_C)
      * ** (Subject Attributes):** Standard physical, social, or functional properties of S.
      * ** (Carrier Latent Properties):** The specific "Achilles' heel," hidden rule, or literal base of C.
      #### 6. Violation / Conflict Points (V)
      * **Category Misalignment:** Treating a person as an object, or a myth as a biological reality.
      * **Functional Interference:** S negates C?s core identity.
      * **Physical Inconsistency:** Scaling, gravity, or biological laws from S's domain overriding C's domain.
      #### 7. Emergent Meaning (I - Implication)
      The "Aha!" moment. How does the "taming" or "disruption" of C by S prove the extreme power, quality, or essence of S?
      ### UNIVERSAL SCHEMA GRAMMAR
      Based on the above analysis, given **{subject}**, derive and generate the corresponding creative image schema Grammar. Output the following elements:
      * **Subject (S):**
      * **Carrier (C):**
      * **Generic Space (G):**
      * **Aesthetic (Aes):**
      * **Composition (Cp):**
      * **Aesthetic Tonality (At):**
      * **Graphic/Typographic Elements (Te):**
      * **Inherent Attributes (A_\sub{S}):**
      * **Carrier Latent Properties (A_C):**
      * **Violation / Conflict Points (V):**
      * **Emergent Meaning (I):**
      After completing the above element analysis, please combine these elements appropriately to generate the final creative image scheme for **{subject}**. This scheme should clearly define the main subject, metaphorical carriers, visual conflict, composition method, aesthetic style, textual elements, and the final creative meaning conveyed, making it directly usable for image generation or advertising creative design.

    \end{lstlisting}
\captionof{figure}{System prompt for strong prompt ablation study}
\label{fig:strong_prompt}

\section{User Study Settings and Questions}
The definition of task 1 is as follows:
(1) Metaphor Recognizability (Recogn.): whether the metaphorical relationship is immediately perceivable (1=completely unclear, 5=instantly recognizable); (2) Metaphor Ingenuity: how creative and clever the metaphorical design is (1=mundane, 5=exceptionally creative); (3) Violation Appropriateness (Violation): whether the category violations appear meaningful rather than arbitrary (1=jarring/nonsensical, 5=purposeful and apt); (4) Visual Integration: how naturally the subject blends with the carrier (1=crude collage, 5=seamless fusion); and (5) Overall Visual Quality (Visual): the overall quality of composition, lighting, and details (1=poor quality, 5=professional standard). 

We also present user study question in SFig.~\ref{fig:user_study_question}
\begin{lstlisting}
Metaphor Recognition (MR): ``Can you immediately identify what object/concept is used as the metaphor?'' (1=completely unclear, 5=instantly obvious)
Metaphor Aptness (MA): ``Is the metaphor appropriate and clever for promoting [target product]?'' (1=far-fetched, 5=brilliant)
Conceptual Integration (CI): ``Does the fusion of product and metaphor vehicle appear natural?'' (1=awkward collage, 5=seamless blend)
Visual Appeal (VA): ``How visually attractive is this advertisement?'' (1=unappealing, 5=highly attractive)
Message Clarity (MC): ``After viewing the ad, are you clear about the product's core benefit?'' (1=very unclear, 5=crystal clear)
Overall Creative Quality (OCQ): ``How would you rate the overall creative quality?" (1=mediocre, 5=exceptional)'' (1=very unclear, 5=crystal clear)
    \end{lstlisting}
\captionof{figure}{User study questions.}
\label{fig:user_study_question}

\section{Applications}
\paragraph{Commercial product advertisements}
In the realm of commercial advertising, our framework facilitates the automated synthesis of high-impact visual metaphors by mapping product attributes onto novel creative carriers. The system provides significant versatility, supporting both text-based descriptions and image-based references as the driving subjects for promotional design Fig.~8 (main text). By precisely inducing ``Category Violations'' to stimulate ``Emergent Meaning'', our approach ensures high-fidelity visual results while establishing an end-to-end pipeline for the efficient production of narrative-driven marketing content across various digital platforms.

\paragraph{Meme generation}
The proposed framework demonstrates significant potential in the automated generation of internet memes, a creative domain where communicative impact and humor are deeply rooted in profound visual metaphors and cognitive dissonance, as shown in Fig.~9 (main text). By precisely extracting the ``Generic Space'' from canonical meme templates, our approach facilitates the seamless transfer of underlying logical mechanisms and satirical intent to emerging specific target entities while preserving the structural integrity of the original metaphorical framework. Consequently, the system achieves nuanced ``Category Violations'' that enhance both visual wit and compositional coherence, establishing a robust technical pipeline for high-fidelity, context-aware content synthesis and personalized expression in digital social media.

\section{Bad Case Ayalysis}
The bad case of our method lies in the excessive cognitive barrier required to decode certain migrated metaphors, which may hinder instantaneous communication.
{As shown at the bottom of Fig.~10 in the \emph{main text} (not a supplementary figure),} the ``Achilles' Heel'' allusion for a band-aid advertisement relies heavily on the viewer's specific cultural background.
Without recognizing the mythological context, the symbolic ``ultimate protection'' is reduced to a literal historical injury, losing its persuasive power.
Similarly, the ``starved Siren'' metaphor for noise-canceling headphones necessitates an exhaustive multi-step logical inference (Siren's song $\rightarrow$ attraction $\rightarrow$ predation $\rightarrow$ noise-blocking $\rightarrow$ starvation).
These instances highlight the trade-off between semantic depth and cognitive immediacy. In such cases, the agentic reasoning may prioritize logical completeness over ease of interpretation, yielding increase in the viewer's decoding effort.

\section{More Results}
\begin{figure*}[!h]
  \centering
  \includegraphics[width=1\linewidth]{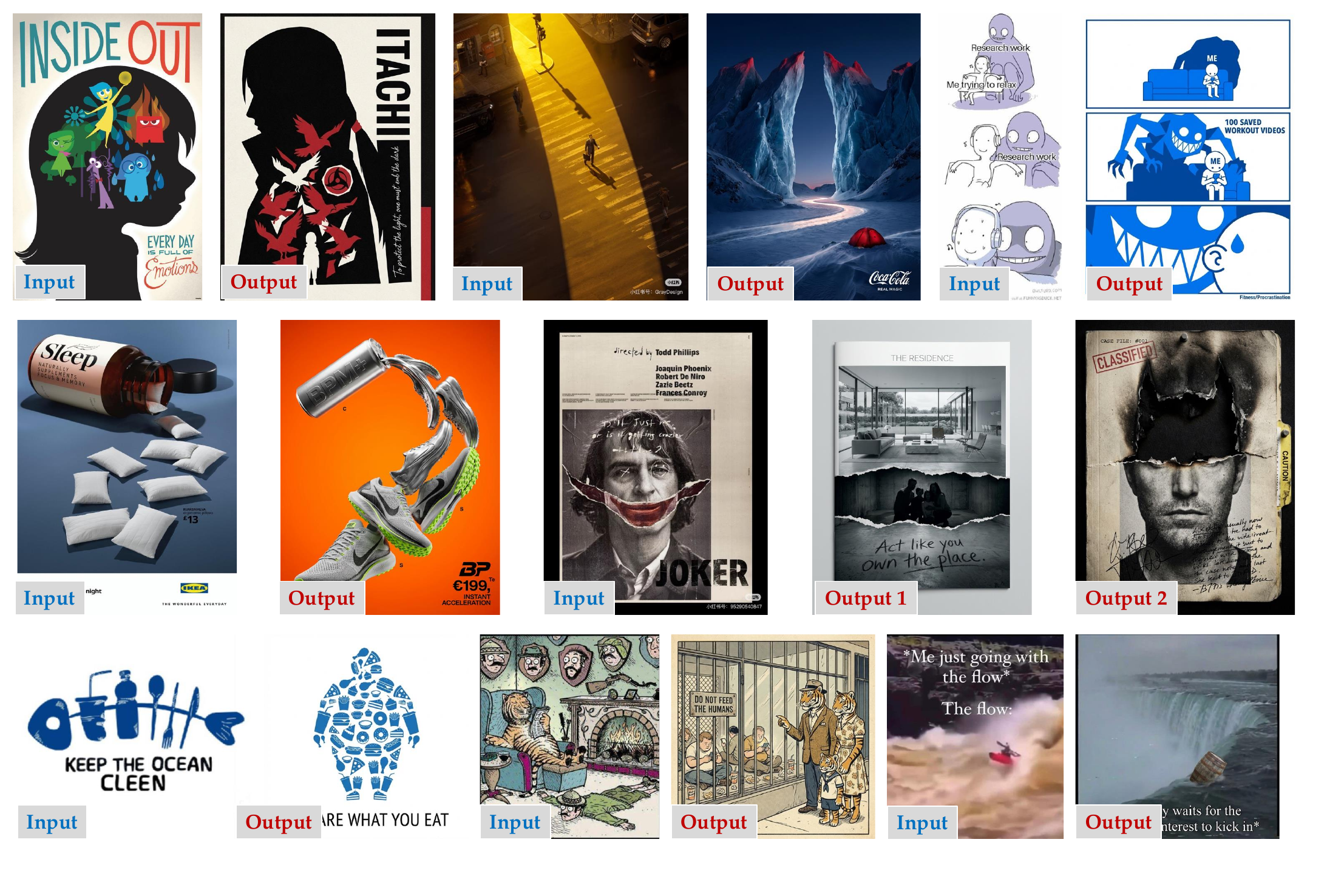}
  \caption{More results.}
  \label{fig:more_res}
\end{figure*}

We present more results in SFig.~\ref{fig:more_res}

\clearpage